\documentclass[twoside,11pt]{article}

\usepackage{jmlr2e}
\usepackage{amssymb}
\usepackage{amsmath}
\usepackage{subfig}
\usepackage{float}
\usepackage{graphicx}
\RequirePackage[colorlinks,citecolor=blue,urlcolor=blue]{hyperref}
\usepackage{natbib}
\usepackage{cleveref}
\usepackage{tikz}
\usetikzlibrary{fit,positioning,calc}
\usepackage{algorithm}
\usepackage{algorithmic}
\usepackage[utf8]{inputenc}
\usepackage{graphicx}
\usepackage{colortbl,array} 
\usepackage{multirow,bigdelim}
\usepackage{booktabs}

\def\T{{ \mathrm{\scriptscriptstyle T} }}
\makeatletter
\newenvironment{breakablealgorithm}
  {
   \begin{center}
     \refstepcounter{algorithm}
     \hrule height.8pt depth0pt \kern2pt
     \renewcommand{\caption}[2][\relax]{
       {\raggedright\textbf{\ALG@name~\thealgorithm} ##2\par}%
       \ifx\relax##1\relax 
         \addcontentsline{loa}{algorithm}{\protect\numberline{\thealgorithm}##2}%
       \else 
         \addcontentsline{loa}{algorithm}{\protect\numberline{\thealgorithm}##1}%
       \fi
       \kern2pt\hrule\kern2pt
     }
  }{
     \kern2pt\hrule\relax
   \end{center}
  }
\makeatother

\newtheorem{thm}{Theorem}

\newtheorem{prop}[thm]{Proposition}

\ShortHeadings{Bayesian Learning of Dynamic Multilayer Networks}{}
\firstpageno{1}

\begin{document}

\title{Bayesian Learning of Dynamic Multilayer Networks}

\author{\name Daniele Durante \email durante@stat.unipd.it\\
      \addr Department of Statistical Sciences\\
       University of Padova\\
       Padova, 35121, Italy\\
       \AND
\name Nabanita Mukherjee \email nm168@duke.edu \\
 \addr Department of Statistical Science\\
     Duke University\\
       Durham, NC 27708-0251, USA \\
        \AND      
\name Rebecca C. Steorts \email beka@stat.duke.edu \\
      \addr Departments of Statistical Science and Computer Science\\
     Duke University\\
       Durham, NC 27708-0251, USA
}

\editor{}

\maketitle

\begin{abstract}%
A plethora of networks is being collected in a growing number of fields, including disease transmission, international relations, social interactions, and others. As data streams continue to grow, the complexity associated with these highly multidimensional connectivity data presents novel challenges. In this paper, we focus on the time-varying interconnections among a set of actors in multiple contexts, called layers. Current literature lacks flexible statistical models for dynamic multilayer networks, which can enhance quality in inference and prediction by efficiently borrowing information within each network, across time, and between layers. Motivated by this gap, we develop a Bayesian nonparametric model leveraging latent space representations. Our formulation characterizes the edge probabilities as a function of shared and layer-specific actors positions in a latent space, with these positions changing in time via Gaussian processes. This representation facilitates dimensionality reduction and incorporates different sources of information in the observed  data. In addition, we obtain tractable procedures for posterior computation, inference, and prediction. We provide theoretical results on the flexibility of our model. Our methods are tested on simulations and infection studies monitoring dynamic face-to-face contacts among individuals in multiple days, where we perform better than current methods in inference and prediction.
\end{abstract}


\section{Introduction}
\label{s:intro}
Data on social interaction processes are rapidly becoming highly multidimensional, increasingly complex and inherently dynamic, providing a natural venue for machine learning as seen in applications such as disease transmission \citep{mar_2013}, international relations \citep{she_2016}, social interactions \citep{guo_2015}, and others. While modeling of a single network observation is still an active area of research, the increasing availability of multidimensional networks from World Wide Web architectures \citep{leet_2013}, telecommunication infrastructures \citep{blondel_2015} and human sensing devices \citep{cat_2010}, have motivated more flexible statistical models. This endeavor is particularly relevant in the dynamic multilayer network field, providing information on the time-varying connectivity patterns among a set of actors in different contexts. Notable examples include  dynamic relationships between individuals according to multiple forms of social interactions \citep{snij_2013}, and time-varying connectivity structures among countries based on different types of international relations \citep{hoff_2015}. 

In modeling these highly complex networks, it is of paramount interest to learn the wiring processes underlying the observed data and to infer differences in networks'  structures across layers and times. Improved estimation of the data's generating mechanism can refine the understanding of social processes and enhance the quality in prediction of future networks. In order to successfully accomplish these goals, it is important to define  statistical models which can incorporate the different sources of information in the observed data, without affecting flexibility.  However, current literature lacks similar methods, to our knowledge. 

Motivated by this gap, we develop a Bayesian nonparametric model for dynamic multilayer networks which efficiently incorporates dependence within each network, across time and between the different layers, while preserving flexibility. Our formulation borrows network information by defining the edge probabilities as a function of pairwise similarities between actors in a latent space. To share information among layers without affecting flexibility in characterizing layer-specific structures, we force a subset of the latent coordinates of each actor to be common across layers and let the remaining coordinates to vary between layers. Finally, we accommodate network dynamics by allowing these actors' coordinates to change in time, and incorporate time information by modeling the dynamic actors' coordinates via Gaussian processes  \citep[e.g.][]{ras_2006}. Our model is tractable and has a theoretical justification, while providing simple procedures for  inference and prediction. In addition, we find that our procedures out perform current methods in inference and out-of-sample prediction on both simulated and real data.

The paper proceeds as follows. Section \ref{sec:relatedWork} reviews recent contributions in the literature, and Section \ref{s:1dat} provides a motivational example of dynamic multilayer networks.  Section \ref{s:2} describes our Bayesian latent space model, while Section \ref{s:7} discusses its properties. Methods for posterior computation and prediction are provided in Section \ref{s:3}. Section  \ref{s:4} highlights the performance gains of our model compared to other possible competitors in a simulation study, whereas Section  \ref{s:5} illustrates on infection studies monitoring face-to-face contacts our success in inference and prediction compared to possible competitors. Future directions and additional fields of application are discussed in Section \ref{s:6}.

\subsection{Related Work}
\label{sec:relatedWork}
Current literature for multidimensional network data considers settings in which  the multiple networks are either dynamic or multilayer. Statistical modeling of dynamic networks has focused on  stochastic processes which are designed to borrow information between edges and across time  \citep[e.g.][]{hann_2007, snij_2010,xing_2010,yang_2011,dur_2014, crane_2016}, whereas inference for multilayer networks has motivated formulations which can suitably induce dependence between edges and across the different types of relationships --- characterizing the multiple layers \citep[e.g.][]{goll_2016,  han2015,hea_2014}.  These contributions have generalized exponential random graph models \citep{holl_1981,fra_1986} and latent variables models \citep{now_2001,hof_2002,air_2008} for a single network to allow inference in  multidimensional frameworks, when the multiple networks arise either from dynamic or multilayer studies. These methods are valuable building blocks for more flexible models, but fall far short of the goal of providing efficient procedures in more complex settings when the networks are both dynamic and multilayer.

The routine collection of dynamic multilayer networks is a recent development and statistical modeling of such data is still in its infancy. For example, \cite{lee_2011} considered a generalization of exponential random graph models for multilayer networks, but performed a separate analysis for each time point.  \cite{ose_2014} focused instead on a dynamic stochastic block model which borrows information across time and within each network, but forces the underlying block structures to be shared between layers. Dynamic multilayer networks are complex objects combining homogenous structures shared between actors, layers, and smoothly evolving across time, with layer-specific patterns and  across-actor heterogeneity. Due to this, any procedure that fails to incorporate the different sources of information in the observed data \citep[e.g.][]{lee_2011} is expected to lose efficiency, whereas models focusing on shared patterns  \citep[e.g.][]{ose_2014} may lack flexibility. 

More general formulations are the multilayer stochastic actor-oriented model \citep{snij_2013} and the multilinear tensor regression \citep{hoff_2015}. \cite{snij_2013} allowed for dynamic inference on network properties within and between layers, but failed to incorporate across-actor heterogeneity. This may lead to a lack of flexibility in prediction. \cite{hoff_2015} considered autoregressive models with the vector of parameters having a tensor factorization representation. This formulation allows for across-actor heterogeneity, but forces the model parameters to be constant across time. In addition,  the parameterization of the interdependence between layers relies on homogeneity assumptions. Consistent with these methods, our representation incorporates the different types of dependencies in the observed data, but crucially preserves flexibility to avoid restrictive homogeneity assumptions.

\subsection{Motivating Application}
\label{s:1dat}
Our motivation is drawn from epidemiologic studies monitoring hourly face-to-face contacts among individuals in a rural area of Kenya during three consecutive days. Data are available from the human sensing platform SocioPatterns (\url{http://www.sociopatterns.org}) and have been collected using wearable devices that exchange low-power radio packets when two individuals are located within a sufficiently close distance to generate a potential occasion of  contagion. Leveraging this technology it is possible to measure for each hour in the three consecutive days which pairs of actors had a face-to-face proximity contact. These information are fundamental to monitor the spread of diseases and learn future patterns. Refer also to \cite{cat_2010} for a detailed description of the data collection technology and \cite{kiti_2016} for a more in-depth discussion of our motivating application. 
\begin{figure}[t]
\centering
\includegraphics[width=14.2cm]{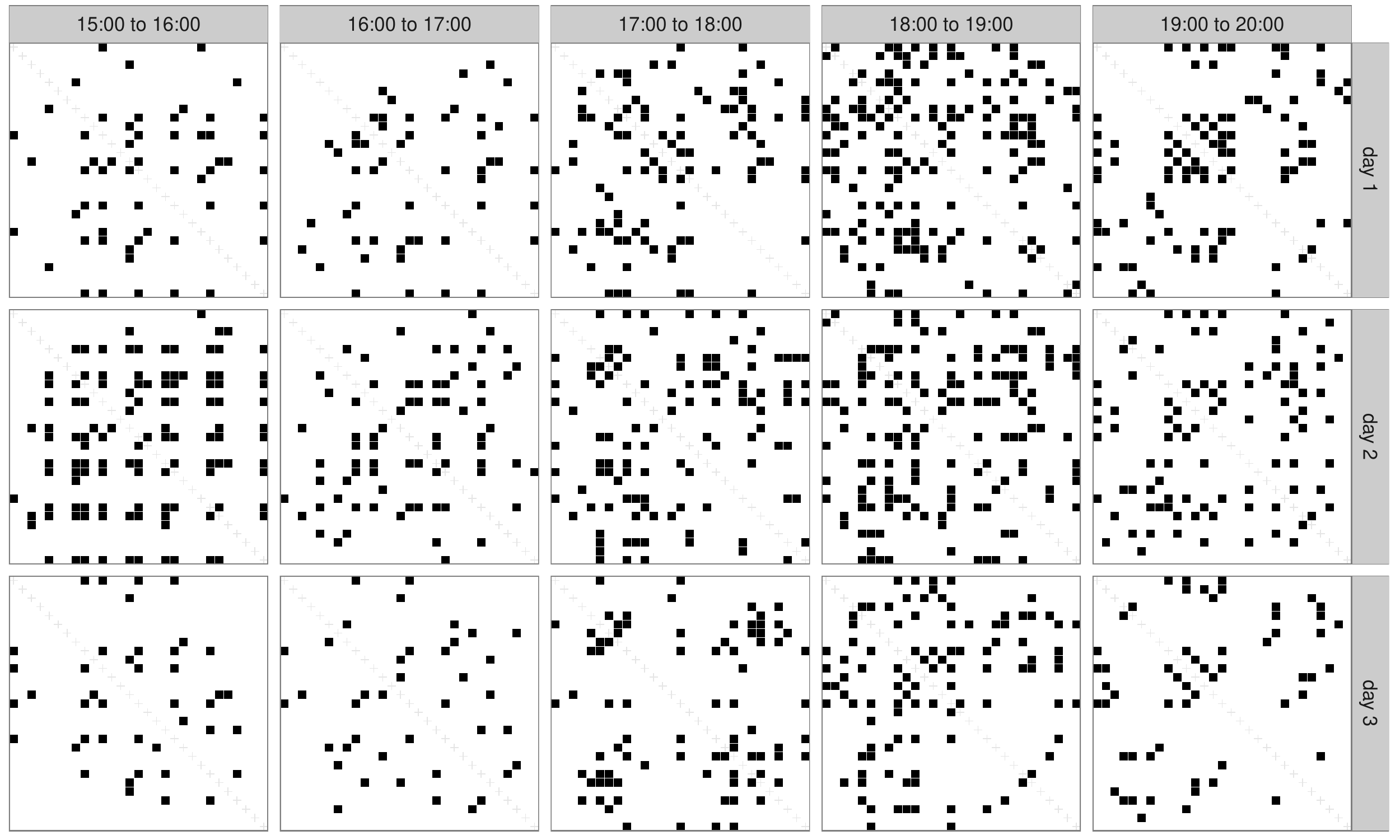}
\caption{For five consecutive hours in the three days, adjacency matrices $Y^{(k)}_{t_i}$ representing the observed face-to-face contact networks. Black refers to an edge and white to a non-edge.}
\label{f1}
\end{figure}

As illustrated in Figure \ref{f1}, the observed data can be structured as a dynamic multilayer network characterized by a sequence of $V\times V$ symmetric adjacency matrices $Y^{(k)}_{t_i}$ collected on a common time grid $t_1, \ldots, t_n$ in different days $k=1, \ldots, K$. The matrices have entries $Y^{(k)}_{t_i[vu]}=Y^{(k)}_{t_i[uv]}=1$ if actors $v$ and $u$ interact at time $t_i$ in day $k$, and  $Y^{(k)}_{t_i[vu]}=Y^{(k)}_{t_i[uv]}=0$ otherwise, for every $v=2, \ldots, V$ and $u=1, \ldots, v-1$. Dynamic multilayer proximity data  provide key information on infections spread \citep{cauc_2011}, facilitating the design of outbreak prevention policies \citep{van_2013}. The importance of this endeavor has motivated a wide variety of descriptive studies summarizing the information provided by  face-to-face proximity data in several environments, including  hospitals  \citep{van_2013}, schools \citep{ste_2011} and households \citep{kiti_2016}. 

Although the above analyses provide key insights on the connectivity processes underlying the observed proximity data, explicitly accounting for variability in network structures via carefully tailored statistical models can lead to improved learning of the connectivity patterns and properties, while providing methodologies for formal inference in the network framework, including estimation techniques, uncertainty quantification and prediction. In Section \ref{s:5}, we apply our methodology to this motivating application, showing how a careful statistical model for dynamic multilayer network data can provide substantial performance gains in learning underlying social processes and predicting future contact patterns.

\section{Dynamic Multilayer Latent Space Model}
\label{s:2}
Let $Y^{(k)}_{t_{i}}$ denote the $V \times V$ adjacency matrix characterizing the undirected network observed at time $t_i \in \Re^{+}$  in layer $k$, for any $t_i=t_1, \ldots, t_n$ and  $k=1, \ldots, K$. Each matrix has binary elements $Y^{(k)}_{t_{i}[vu]}=Y^{(k)}_{t_{i}[uv]} \in \{0,1\}$ measuring the absence or presence of an edge among actors $v$ and $u$ at time $t_i$ in layer $k$, for every $v=2, \ldots, V$ and $u=1, \ldots, v-1$. When modeling the data $Y=\{Y^{(k)}_{t_{i}}: t_i=t_1, \ldots, t_n, \ k=1, \ldots, K \}$, we assume that $Y$ is a realization, on a finite time grid, of the  stochastic process $\mathcal{Y}=\{\mathcal{Y}^{(k)}(t): t \in \mathbb{T} \subset \Re^+, \ k=1, \ldots, K \}$, and seek a provably flexible representation for the probabilistic generative mechanism associated with the stochastic process $\mathcal{Y}$. Note that, as each random adjacency matrix $\mathcal{Y}^{(k)}(t)$ is symmetric and the diagonal elements --- denoting self-relations --- are not of interest, it is sufficient to model its lower triangular elements $\{\mathcal{Y}^{(k)}_{[vu]}(t):  v=2, \ldots, V,\ u=1, \ldots, v-1\}$. 

\begin{table}[t]
\begin{center}
\def~{\hphantom{0}}
{%
\begin{tabular}{l p{10.3cm}}
{\bf Type of information} & {\bf Illustrative example}  \\
\hline
Network information & {\small{If individual $v$ had a face-to-face contact with both actors $u$ and $w$ at time $t_i$ in day $k$, this information may be relevant to learn the face-to-face contact behavior between $u$ and $w$ at time $t_i$ in day $k$.}} \\
\hline
Layer information  & {\small{If individuals $v$ and $u$ had a face-to-face contact at time $t_i$ in day $k$, this information may be relevant to learn the contact behavior between $v$ and $u$ at the same time $t_i$ in other days.}}\\
\hline
Time information  &  {\small{If individuals $v$ and $u$ had a face-to-face contact at time $t_i$ in day $k$, this information may be relevant to learn the contact behavior between $v$ and $u$ at the next time $t_{i+1}$ in the same day.}}\\
 \hline
\end{tabular}}
\caption{Relevant sources of information characterizing a dynamic multilayer network. }
\label{table0}
\end{center}
\end{table}

One major modeling objective is to carefully incorporate dependence among edges, between layers and across time, without affecting flexibility. Recalling the motivating application in Section \ref{s:1dat} and Figure \ref{f1}, it is reasonable to expect three main sources of information in the dynamic multilayer face-to-face contact data, summarized in Table \ref{table0}. Incorporating such information can substantially improve the quality of inference and prediction, while facilitating dimensionality reduction. However, in reducing dimensionality and borrowing of information, it is important to avoid restrictive formulations that lead to inadequate characterizations of dynamic patterns, layer-specific structures, and across-node heterogeneity.

We accomplish the aforementioned goals via a flexible dynamic latent bilinear model combining shared and layer-specific actors coordinates which are allowed to change in time via Gaussian process priors. Section \ref{sub:21} describes our model formulation with a focus on sharing network and layer information, whereas Section \ref{sub:22} clarifies how time information is effectively incorporated via Gaussian processes.

\subsection{Dynamic Bilinear Model with Shared and Layer-Specific Coordinates}
\label{sub:21}
In order to incorporate network and layer information, while preserving flexibility, we adapt latent bilinear models  for a single network \citep{hoff_2005} to our  dynamic multilayer framework. In particular, we characterize the edges as conditionally independent Bernoulli random variables given their corresponding edge probabilities, and borrow information within each network by defining these probabilities as a function of the actors coordinates in a latent space. In order to incorporate dependence among the different layers, we force a subset of these coordinates to be common across layers and allow the remaining coordinates to change between layers. This last choice facilitates a flexible characterization of more irregular events and incorporates differences between layers in specific contact patterns. 

Consistent with the aforementioned statistical model, and  letting $\pi^{(k)}_{[vu]}(t) \in (0,1)$  be the probability of an edge between actors $v$ and $u$ at time $t \in \mathbb{T}$ in layer $k$, we define
\begin{eqnarray}
\mathcal{Y}^{(k)}_{[vu]}(t) \mid \pi^{(k)}_{[vu]}(t) \sim \mbox{Bern}\{\pi^{(k)}_{[vu]}(t)\}, 
\label{eq1}
\end{eqnarray}
independently for every time  $t \in \mathbb{T} $, layer $k=1, \ldots, K$ and pair of nodes $[vu]$, $v=2, \ldots, V$, $u=1, \ldots, v-1$. To incorporate our multilayer bilinear representation having shared and layer-specific latent coordinates, we express the log-odds of each edge probability as 
\begin{eqnarray}
\mbox{logit}\{\pi^{(k)}_{[vu]}(t)\}=\mu(t)+ \sum_{r=1}^{R}\bar{x}_{vr}(t)\bar{x}_{ur}(t)+ \sum_{h=1}^{H}x^{(k)}_{vh}(t)x^{(k)}_{uh}(t),
\label{eq2}
\end{eqnarray}
where $\bar{x}_{vr}(t) \in \Re$ is the $r$th coordinate of actor $v$ at time $t$ shared across the different layers, whereas $x^{(k)}_{vh}(t) \in \Re$ denotes the $h$th coordinate of actor $v$ at time $t$ specific to layer $k$, for every $t \in \mathbb{T}$, $k=1, \ldots, K$, $v=1, \ldots, V$, $r=1, \ldots, R$ and $h=1, \ldots, H$. Finally, $\mu(t) \in \Re$ represents a time-varying baseline parameter centering the log-odds processes to improve computational and mixing performance.

In equation \eqref{eq2} the probability of an edge between actors $v$ and $u$ at time $t \in \mathbb{T}$ in layer $k$, increases with  $\sum_{r=1}^R \bar{x}_{vr}(t)\bar{x}_{ur}(t)$ and $\sum_{h=1}^H x^{(k)}_{vh}(t)x^{(k)}_{uh}(t)$. Note that $\sum_{r=1}^R \bar{x}_{vr}(t)\bar{x}_{ur}(t) \in \Re$ characterizes a similarity measure between actors $v$ and $u$ at time $t$ common to all the layers, whereas $\sum_{h=1}^H x^{(k)}_{vh}(t)x^{(k)}_{uh}(t)\in \Re$ defines a layer-specific deviation from this shared similarity, which enhances flexibility in modeling network structures specific to layer $k$. These similarity measures are defined as the dot product of shared and layer-specific actors coordinates in a latent space, allowing actors with coordinates in the same direction to have an higher chance of contact than actors with coordinates in opposite directions. 

The dot product characterization in equation \eqref{eq2} is in the same spirit as the factorization in \cite{hoff_2005}  for a single network and allows dimensionality reduction from $n\times K\times V \times(V-1)/2$ time-varying log-odds to $n \times\{1+ V \times (R+ H\times K)\}$  dynamic latent processes, where typically $R \ll V$ and $H \ll V$. Although it is possible to consider other concepts of similarity when relating the latent coordinates to the log-odds \citep[e.g.][]{hof_2002}, the dot product representation facilitates the definition of simple algorithms for posterior inference and has been shown to efficiently borrow information within the network while accommodating topological structures of interest such as homophily, transitivity, and others. 

\subsubsection{Interpretation and Identifiability}
Recall the motivating application described in Section \ref{s:1dat}, our representation has an intuitive interpretation. In fact, the latent coordinates of each actor may represent his propensity towards $R+H$ latent interests or tasks. According to the factorization in equation \eqref{eq2}, actors having propensities for the different interests in the same directions are expected to be more similar than actors with interests in opposite directions.  Therefore, these actors are more likely to interact according to equation \eqref{eq1}. Part of these actors' interests or tasks  may be associated with a common routine or constrained by daytime schedules. Therefore, the actors propensities towards these interests are expected to remain constant across days providing a subset of $R$ shared coordinates. Finally, these propensities are allowed to change across time inducing dynamic variations in the contacts within each day.

Although inference for the latent coordinates is potentially of interest, the factorization in equation \eqref{eq2} is not unique. In fact, it is easy to show that there exist many different latent coordinate values leading to the same collection of edge probabilities under the decomposition in equation \eqref{eq2}. However, such an over-complete representation does not lead to identifiability issues if inference focuses on identified functionals of the latent coordinates and has beneficial effects in terms of posterior computation and borrowing of information \citep{bhatt_2011,ghosh_2009}. Our focus is not on estimating the latent coordinates but on leveraging the factorization in equation \eqref{eq2} to enhance efficiency in learning of identified functionals of the edge probabilities and improve prediction of future networks. Therefore, we avoid identifiability constraints on the latent coordinates in equation \eqref{eq2} as they are not required to ensure identifiability of the edge probabilities.

\subsection{Gaussian Process Priors for the Time-Varying Latent Coordinates}
\label{sub:22}
Equation \eqref{eq2} facilitates borrowing information among edges and between layers. This is obtained by  leveraging the shared dependence on a common set of latent coordinates. We additionally incorporate across-layer heterogeneity by utilizing a set of layer-specific latent positions. Moreover, these coordinates are allowed to vary across time in order to accommodate dynamic variations in the multilayer network structure. While it is possible to estimate the edge probabilities separately for each time point, this approach may be suboptimal in ruling out temporal dependence. As seen in Figure \ref{f1}, it is reasonable to expect a degree of dependence between contact networks observed at close times, with this dependence decreasing with the time lag. Therefore, a formulation sharing information across time may provide substantial benefits in terms of efficiency, uncertainty propagation, and prediction.

Motivated by dynamic modeling of a single network, \cite{dur_2014} addressed a related goal by considering Gaussian process priors for the actors' latent coordinates with an additional shrinkage effect  to facilitate automatic adaptation of the latent space dimensions and avoid overfitting. In the spirit of their proposed methods, we define
\begin{eqnarray}
\bar{x}_{vr}(\cdot) \sim \mbox{GP}(0,\tau^{-1}_{r}c_{\bar{x}}), \quad \ \ \ \mbox{with  } \ c_{\bar{x}}(t_i,t_j)=\mbox{exp}\{-\kappa_{\bar{x}}(t_i-t_j)^2 \}, \kappa_{\bar{x}}>0,
\label{eq3}
\end{eqnarray}
independently for $v=1, \ldots, V$, $r=1, \ldots, R$ and 
\begin{eqnarray}
 {x}^{(k)}_{vh}(\cdot) \sim \mbox{GP}(0,\tau^{(k)-1}_{h}c_{{x}}), \quad \mbox{with  } \ c_{x}(t_i,t_j)=\mbox{exp}\{-\kappa_{{x}}(t_i-t_j)^2 \}, \kappa_{{x}}>0,
 \label{eq4}
\end{eqnarray}
independently for $v=1, \ldots, V$, $h=1, \ldots, H$ and $k=1, \ldots, K$. In the prior specification given in equations \eqref{eq3}--\eqref{eq4}, the quantities $c_{\bar{x}}(t_i,t_j)$ and $c_{x}(t_i,t_j)$ denote the squared exponential correlation functions of the Gaussian processes for the shared and layer-specific latent coordinates, respectively. The quantities $\tau^{-1}_{1}, \ldots, \tau^{-1}_{R}$ and $\tau^{(k)-1}_{1}, \ldots ,\tau^{(k)-1}_{H}$, for each $k=1, \ldots, K$, are instead positive shrinkage parameters controlling the concentration of the latent coordinates around the zero constant mean function. Focusing on the prior for the trajectories of the $r$th shared coordinates $\bar{x}_{vr}(\cdot)$, $v=1, \ldots, V$,  a value of $\tau^{-1}_{r}$ close to zero forces these trajectories to be concentrated around the zero constant mean function, reducing the effect of the $r$th shared dimension in defining the edge probabilities in equation~\eqref{eq2}.

In order to penalize over-parameterized representations and facilitate adaptive deletion of unnecessary dimensions, the above shrinkage effects are designed to be increasingly strong as the latent coordinates indices $r=1, \ldots, R$ and $h=1, \ldots, H$ increase. This goal is accomplished by considering the following multiplicative inverse gamma priors  \citep{bhatt_2011} for the shrinkage parameters:
\begin{eqnarray}
\frac{1}{\tau_{r}} &=& \prod_{m=1}^r \frac{1}{\delta_m}, \  \ r=1, \ldots, R,  \ \mbox{with} \ \delta_1 \sim \mbox{Ga}(a_1,1), \ \delta_{m \geq 2} \sim \mbox{Ga}(a_2,1),  \label{eq5}\\
\frac{1}{\tau^{(k)}_{h}} &=& \prod_{l=1}^h \frac{1}{\delta^{(k)}_l}, \  \ h=1, \ldots, H, \ k=1, \ldots, K,  \ \mbox{with} \ \delta^{(k)}_1 \sim \mbox{Ga}(a_1,1), \ \delta^{(k)}_{l \geq 2} \sim \mbox{Ga}(a_2,1). \ \ \ \ \ \label{eq6}
\end{eqnarray}
According to \cite{bhatt_2011}, the multiplicative inverse gamma prior in equations \eqref{eq5}--\eqref{eq6} induces prior distributions for the shrinkage parameters with a cumulative shrinkage effect. In particular, these priors are increasingly concentrated close to zero as the indices $r=1, \ldots, R$ and $h=1, \ldots, H$ increase, for appropriate $a_1\in \Re^+$ and $a_2\in \Re^+$, facilitating adaptive dimensionality reduction and reducing over-fitting issues. Refer also to \cite{durante_MGP} for an additional discussion on the multiplicative inverse gamma prior and its shrinkage properties. To conclude the prior specification, we let $\mu(\cdot) \sim \mbox{GP}(0, c_{\mu})$, with $c_{\mu}(t_i,t_j)=\mbox{exp}\{-\kappa_{\mu}(t_i-t_j)^2 \}$, $\kappa_{\mu}>0$. 

The Gaussian process prior provides an accurate choice in our setting which incorporates time dependence, allowing the amount of information shared between networks to increase as the time lag decreases. Beside improving efficiency, the Gaussian process can deal with multilayer networks observed at potentially unequally spaced time grids and is closely related to the multivariate Gaussian random variable, providing substantial benefits in terms of computational tractability and interpretability. In fact, following \cite{ras_2006}, equations \eqref{eq3}--\eqref{eq4} imply the following prior for the shared and layer-specific latent coordinates at the finite time grid $t_1, \ldots, t_n$ on which data are observed:
\begin{eqnarray}
\{\bar{x}_{vr}(t_1),\ldots, \bar{x}_{vr}(t_n)\}^{\T}\sim {}\mbox{N}_n(0,\tau^{-1}_{r}\Sigma_{\bar{x}}), \
\label{eq7}
\end{eqnarray}
independently for $v=1, \ldots, V$, $r=1, \ldots, R$ and
\begin{eqnarray}
\{{x}^{(k)}_{vh}(t_1),\ldots, {x}^{(k)}_{vh}(t_n)\}^{\T}{} \sim {}\mbox{N}_n(0,\tau^{(k)-1}_{h}\Sigma_{{x}}),    
\label{eq8}
\end{eqnarray}
independently for $v=1, \ldots, V$,  $h=1, \ldots, H$,  and  $k=1, \ldots, K$. In equations \eqref{eq7}--\eqref{eq8}, the $n \times n$ variance and covariance matrices  $\Sigma_{\bar{x}}$ and $\Sigma_{{x}}$ have elements $\Sigma_{\bar{x}[ij]}=\mbox{exp}\{-\kappa_{\bar{x}}(t_i-t_j)^2 \}$ and $\Sigma_{{x}[ij]}=\mbox{exp}\{-\kappa_{{x}}(t_i-t_j)^2 \}$, respectively. The same holds for the baseline process obtaining $\{\mu(t_1), \ldots, \mu(t_n) \}^{\T}\sim \mbox{N}_n(0,\Sigma_{\mu})$.

\section{Model Properties}
\label{s:7}
In order to ensure accurate learning and prediction, it is important to guarantee that our factorization in equation \eqref{eq2} along with the Gaussian process priors for its components are sufficiently flexible to approximate a broad variety of dynamic multilayer edge probability processes. These properties are stated in Propositions \ref{prop1} and \ref{prop2}. See Appendix A for detailed proofs. In particular, Proposition \ref{prop1} guarantees that equation \eqref{eq2} is sufficiently flexible to characterize any collection $\pi(t)=\{ \pi^{(k)}_{[vu]}(t) \in (0,1): k=1, \ldots, K,  \ v=2, \ldots, V,\ u=1, \ldots, v-1\}$ for every time $t \in  \mathbb{T}$.
\begin{prop}
Let $\pi(t)=\{ \pi^{(k)}_{[vu]}(t) \in (0,1): k=1, \ldots, K, \ v=2, \ldots, V,\ u=1, \ldots, v-1\}$ denote the collection of edge probabilities between the actors in each layer at time $t \in  \mathbb{T}$. Then every collection $\pi(t)$ can be represented as in equation \eqref{eq2} for some $R$ and $H$.
\label{prop1}
\end{prop}

Although Proposition \ref{prop1} guarantees that factorization in equation \eqref{eq2} is sufficiently general, it is important to ensure that the same flexibility is maintained when defining priors on the parameters in equation \eqref{eq2}. Proposition \ref{prop2} guarantees that our model and prior choices induce a prior on the  dynamic multilayer edge probability process with full support.
\begin{prop}
If $ \mathbb{T}$ is compact, then for every $\pi^{0}=\{\pi^{0(k)}_{[vu]}(t) \in (0,1): t \in \mathbb{T}, \ k=1, \ldots, K, \ v=2, \ldots, V,\ u=1, \ldots, v-1\}$ and $\epsilon>0$,
$$\mathrm{pr}\left(\mathrm{sup}_{t\in \mathbb{T}} \left[\sum_{k=1}^{K}\sqrt{ \sum_{v=2}^V \sum_{u=1}^{v-1} \{ \pi^{(k)}_{[vu]}(t) -\pi^{0(k)}_{[vu]}(t) \}^2}\right]<\epsilon \right)>0,$$
with $\pi^{(k)}_{[vu]}(t)$ factorized as in equation \eqref{eq2} with Gaussian process priors on the latent coordinates.
\label{prop2}
\end{prop}
Full prior support is a key property in ensuring that our Bayesian formulation assigns a positive probability to a neighborhood of every possible true dynamic multilayer edge probability process, avoiding flexibility issues that may arise if zero mass is assigned to a subset of the parameters space where the truth may be.

In order to highlight the dependence structures induced by our model and priors, we additionally study the prior variances and covariances associated with the log-odds processes $z=\{z^{(k)}_{[vu]}(t)=\mbox{logit}\{\pi^{(k)}_{[vu]}(t)\} \in \Re: t \in \mathbb{T}, \ k=1, \ldots, K,  \ v=2, \ldots, V,\ u=1, \ldots, v-1\}$. By conditioning on the shrinkage parameters, and leveraging equations \eqref{eq7} and \eqref{eq8}, we obtain
$$\mathrm{var}  \{z^{(k)}_{[vu]}(t) \mid \tau,\tau^{(k)}\}=1+\sum_{r=1}^{R}\tau_{r}^{-2}+\sum_{h=1}^{H}\tau_h^{(k)-2}, \quad \mathrm{cov} \{z^{(k)}_{[vu]}(t), z^{(k)}_{[pq]}(t) \mid\tau,\tau^{(k)}\}=1,$$
for every layer $k=1, \ldots, K$, time $t \in \mathbb{T}$ and pairs of actors $[vu]$, $v>u$ and $[pq]$, $p>q$ with $p \neq v$ or $q \neq u$. The covariance between different layers, at the same time $t \in \mathbb{T}$ is
$$\mathrm{cov} \{z^{(k)}_{[vu]}(t), z^{(g)}_{[vu]}(t) \mid\tau,\tau^{(k)},\tau^{(g)}\}=1+\sum_{r=1}^{R}\tau_{r}^{-2}, \quad \mathrm{cov} \{z^{(k)}_{[vu]}(t), z^{(g)}_{[pq]}(t) \mid\tau,\tau^{(k)},\tau^{(g)}\}=1,$$
for every pair of layers $k$ and $g$ with $k \neq g$, time $t \in \mathbb{T}$ and pairs of actors $[vu]$, $v>u$ and $[pq]$, $p>q$ with $p \neq v$ or $q \neq u$. Finally, the covariances at different times are
\begin{eqnarray*}
\mathrm{cov}\{z^{(k)}_{[vu]}(t_i), z^{(k)}_{[vu]}(t_j) \mid \tau,\tau^{(k)}\}=e^{ -\kappa_{\mu}(t_i-t_j)^2}+\sum_{r=1}^{R}\tau_{r}^{-2}e^{-2\kappa_{\bar{x}}(t_i-t_j)^2}+\sum_{h=1}^{H}\tau_{h}^{(k)-2}e^{-2\kappa_{{x}}(t_i-t_j)^2}, \\
\mathrm{cov}\{z^{(k)}_{[vu]}(t_i), z^{(g)}_{[vu]}(t_j) \mid \tau,\tau^{(k)},\tau^{(g)}\}=e^{ -\kappa_{\mu}(t_i-t_j)^2}+\sum_{r=1}^{R}\tau_{r}^{-2}e^{-2\kappa_{\bar{x}}(t_i-t_j)^2}, \ \ \ \ \ \ \ \ \ \ \ \ \ \ \ \ \ \ \ \ \ \ \ \ \ \ \\
\mathrm{cov}\{z^{(k)}_{[vu]}(t_i), z^{(k)}_{[pq]}(t_j) \mid \tau,\tau^{(k)}\}=\mathrm{cov}\{z^{(k)}_{[vu]}(t_i), z^{(g)}_{[pq]}(t_j) \mid \tau,\tau^{(k)},\tau^{(g)}\}=e^{ -\kappa_{\mu}(t_i-t_j)^2}, \ \ \ \ \ \  \ \ \ \ \ \ \
\end{eqnarray*}
for every pair of layers $k$ and $g$ with $k \neq g$, times $t_i \in \mathbb{T}$, $t_j\in \mathbb{T}$ and pairs of actors $[vu]$, $v>u$ and $[pq]$, $p>q$ with $p \neq v$ or $q \neq u$.

According to the above results, a priori the log-odds of the edge probabilities have mean zero, whereas their variances increase with the sum of the shared and layer-specific shrinkage parameters. Allowing a subset of latent coordinates to be common to all layers has the effect of introducing dependence between the log-odds of edge probabilities in different layers. Note also that the  Gaussian process priors provide an efficient choice to incorporate dependence between edge probabilities in different times. The strength of this dependence is regulated by the time lag and the positive smoothing parameters $\kappa_{\mu}$, $\kappa_{\bar{x}}$ and $\kappa_{{x}}$. The lower these quantities, the stronger the dependence is between coordinates in different times.

\section{Posterior Computation and Prediction}
\label{s:3}
Equations \eqref{eq1}--\eqref{eq2} along with the Gaussian process priors for the latent coordinates can be seen as a non-linear Bayesian logistic regression with the parameters entering in a bilinear form. Although posterior computation in this setting is apparently a cumbersome task, leveraging the P\'olya-gamma data augmentation for Bayesian logistic regression \citep{pol_2013} and adapting derivations in \cite{dur_2014}, it is possible to derive a simple and tractable Gibbs sampler having conjugate full conditionals. In fact, the P\'olya-gamma data augmentation allows recasting the problem from a logistic regression to a multiple linear regression having transformed Gaussian response data, while derivations in  \cite{dur_2014} provide results to linearize the factorization in equation \eqref{eq2}. 

Joining the above procedures and exploiting equations \eqref{eq7}--\eqref{eq8}, the updating of the latent coordinates processes at every step simply relies on standard Bayesian linear regression. Algorithm \ref{Algorithm_1} in Appendix B provides derivations and guidelines for step-by-step implementation of our Gibbs sampler. We summarize below the main steps of the MCMC routine.

\begin{description}
\item{{\bf{Step [1]}}: For each time $t_i=t_1, \ldots, t_n$, layer $k=1, \ldots, K$ and pair of nodes $[vu]$, $v=2, \ldots, V$, $u=1, \ldots, v-1$, sample the corresponding P\'olya-gamma augmented data.}
\item{{\bf Step [2]}: Update the baseline process  $\mu=\{\mu(t_1), \ldots, \mu(t_n) \}^{\T}$ from its full conditional multivariate Gaussian distribution. This  is obtained by recasting the logistic regression for $\mu$ in terms of a multiple linear regression having transformed Gaussian response.}
\item{{\bf Step [3]}: Update the shared coordinates. In performing this step we block-sample in turn the coordinates' trajectories $\bar{x}_{(v)}=\{\bar{x}_{v1}(t_1),\ldots,\bar{x}_{v1}(t_n), \dots,\bar{x}_{vR}(t_1),\ldots,\bar{x}_{vR}(t_n) \}^{\T}$ of each actor $v=1, \ldots, V$ conditionally on the others $\{\bar{x}_{(u)}: u \neq v\}$. This choice allows us to linearize equation \eqref{eq2}, with $\bar{x}_{(v)}$ acting as coefficients vector and $\{\bar{x}_{(u)}: u \neq v\}$ representing appropriately selected regressors. Leveraging the P\'olya-gamma data augmentation also this step relies on a multiple linear regression with transformed Gaussian response data, providing Gaussian full conditionals for each  $\bar{x}_{(v)}$, $v=1, \ldots, V$.}
\item{{\bf Step [4]} Update the layer-specific coordinates. For each layer $k=1, \ldots, K$, this step relies on the same strategy considered for the shared coordinates, providing again  Gaussian full conditionals for each ${x}^{(k)}_{(v)}=\{{x}^{(k)}_{v1}(t_1),\ldots,{x}^{(k)}_{v1}(t_n), \dots,{x}^{(k)}_{vH}(t_1),\ldots,{x}^{(k)}_{vH}(t_n) \}^{\T}$, $v=1, \ldots, V$ and $k=1, \ldots, K$. Moreover the updating can be performed separately for each layer, allowing this step to be efficiently implemented in parallel. }
\item{{\bf Step [5] and [6]} The updating of the gamma parameters characterizing the priors in equations \eqref{eq5}--\eqref{eq6} follows  conjugate analysis, proving gamma full conditionals.}
\item{{\bf Step [7]} Update the dynamic multilayer edge probabilities simply by applying equation \eqref{eq2} to the samples of the baseline process, the shared and the layer-specific coordinates.}
\end{description}

In performing posterior computation we set $R$ and $H$ at conservative upper bounds, allowing the multiplicative inverse gamma priors for the shrinkage parameters to delete redundant  latent space dimensions not required to characterize the data. Hence, posterior inference is not substantially affected by the choice of $R$ and $H$, unless these bounds are fixed at excessively low values compared to the complexity of the data analyzed. We additionally assess the goodness of these bounds via in-sample and out-of-sample predictive performance.  

\subsection{Edge Prediction}
\label{sub:3}
Edge prediction is an important topic in dynamic modeling of multilayer networks. For example some networks may have unobserved edges due to inability to monitor certain types of relationships or actors at specific times. Likewise, some layers may be available in time earlier than others, facilitating prediction of those yet missing. The availability of efficient procedures that are able to reconstruct partially unobserved connectivity structures or forecast future networks can have important consequences in many applications, such as destabilization of terrorists networks or  epidemic prevention \citep[e.g.][]{tan_2016}. 

Our statistical model for dynamic multilayer networks in equations \eqref{eq1}--\eqref{eq2} facilitates the definition of simple procedures for formal edge prediction relying on the expectation $\mbox{E}\{\mathcal{Y}^{(k)}_{[vu]}(t_i) \mid Y_{\mbox{\footnotesize{obs}}}\}$ of the posterior predictive distribution for  $\mathcal{Y}^{(k)}_{[vu]}(t_i)$, with $Y_{\mbox{\footnotesize{obs}}}$ denoting the observed data. In fact, under equations \eqref{eq1}--\eqref{eq2}, this functional is easily available as
\begin{eqnarray}
\mbox{E}\{\mathcal{Y}^{(k)}_{[vu]}(t_i) \mid Y_{\mbox{\footnotesize{obs}}}\}=\mbox{E}_{ \pi^{(k)}_{[vu]}(t_i)}(\mbox{E}_{\mathcal{Y}^{(k)}_{[vu]}(t_i)}[\{\mathcal{Y}^{(k)}_{[vu]}(t_i)\mid \pi^{(k)}_{[vu]}(t_i)\} \mid Y_{\mbox{\footnotesize{obs}}}])=\mbox{E}\{\pi^{(k)}_{[vu]}(t_i) \mid Y_{\mbox{\footnotesize{obs}}}\},
\label{eq9}
\end{eqnarray}
for every time $t_i \in \mathbb{T}$, layer $k=1, \ldots, K$ and  actors $v=2, \ldots, V$, $u=1, \ldots, v-1$, where $\mbox{E}\{\pi^{(k)}_{[vu]}(t_i) \mid Y_{\mbox{\footnotesize{obs}}}\}$ simply coincides with the posterior mean of $\pi^{(k)}_{[vu]}(t_i)$. Hence prediction requires the posterior distribution of the edge probabilities. These quantities are available also for unobserved edges by adding the following data augmentation step in Algorithm \ref{Algorithm_1}:
\begin{description}
\item{{\bf Step [8]:} Impute the missing edges given the current state of $\pi^{(k)}_{[vu]}(t_i)$ from a $\mbox{Bern}\{\pi^{(k)}_{[vu]}(t_i)\},$ for all the combinations of times $t_i \in \mathbb{T}$, layers $k=1, \ldots, K$ and actors $v=2, \ldots, V$, $u=1, \ldots, v-1$ corresponding to unobserved edges.}
\end{description}

\section{Simulation Study}
\label{s:4}
We consider a simulation study to evaluate the performance of our methodology in a simple scenario partially mimicking the size and structure of the face-to-face dynamic multilayer networks. Consistent with our aim we assess performance in the tasks outlined in Table \ref{tablesimu}. \begin{table}[t]
\begin{center}
\def~{\hphantom{0}}
{%
\begin{tabular}{l p{11.5cm}}
{\bf Quantity} & {\bf Focus of the performance assessment}  \\
\hline
 {{Edge probabilities}}& {\small{Assess accuracy in learning the true edge probability process underlying the simulated network data. As already discussed, accurate modeling of the edge probabilities $\pi^{(k)}_{[vu]}(t_i)$ is fundamental for inference and prediction.}} \\
\hline
 {{Unobserved edges}} &  {\small{Assess out-of-sample predictive performance in forecasting future unobserved networks leveraging the expectation $\mbox{E}\{\mathcal{Y}^{(k)}_{[vu]}(t_i) \mid Y_{\mbox{\footnotesize{obs}}}\}$ of the posterior predictive distribution for the unobserved edges.}}\\
 \hline
 {{Network density}} & {\small{Accuracy in modeling the expected overall chance of connectivity  $\theta^{(k)}(t_i)=\mbox{E}[\sum_{v=2}^V\sum_{u=1}^{v-1}\mathcal{Y}^{(k)}_{[vu]}(t_i)/\{V(V-1)/2\}]=\sum_{v=2}^V\sum_{u=1}^{v-1}\pi^{(k)}_{[vu]}(t_i)/\{V(V-1)/2\},$ for every time $t_i \in \mathbb{T}$ and layer $k=1, \ldots, K$.}}\\
\hline
 {{Actors degrees}} &  {\small{Accuracy in learning the total number of different individuals $\mbox{d}^{(k)}_{v}(t_i)=\mbox{E}\{\sum_{u\neq v}\mathcal{Y}^{(k)}_{[vu]}(t_i)\}=\sum_{u\neq v}\pi^{(k)}_{[vu]}(t_i),$ each actor $v$ is expected to interact with, for every $v=1, \ldots, V$, time $t_i \in \mathbb{T}$ and layer $k=1, \ldots, K$.}}\\
 \hline
\end{tabular}}
\caption{Relevant quantities of interest for inference, which are considered in performance assessments.  Although it is possible to provide inference on several network properties, from an epidemiological perspective, the network density and actors degrees are of interest in characterizing the overall risk of contagion and the infectivity of each actor, respectively.}
\label{tablesimu}
\end{center}
\end{table}

To highlight the benefits of our methods (Joint Analysis), we  compare the performance  with two  competitors. The first (Collapsed Analysis) assumes the layers arise from a common latent space model, forcing all the actors coordinates to be shared among layers. This formulation is a special case of our model holding out in  \eqref{eq2} the layer-specific similarities. Hence, posterior analysis requires minor modifications of Algorithm \ref{Algorithm_1}. The second competitor (Separate Analysis) estimates the dynamic latent space model of \cite{dur_2014} separately for each layer. Although both methods are reasonable choices, modeling shared structures lacks flexibility in capturing layer-specific patterns, whereas separate analyses may lose efficiency in inference and prediction when the layers share common patterns. We describe the simulation setting in Section \ref{s51} and discuss results in Section \ref{s52}.

\subsection{Outline of the Simulation Settings}
\label{s51}
\begin{figure}[t]
\centering
\includegraphics[width=14.9cm]{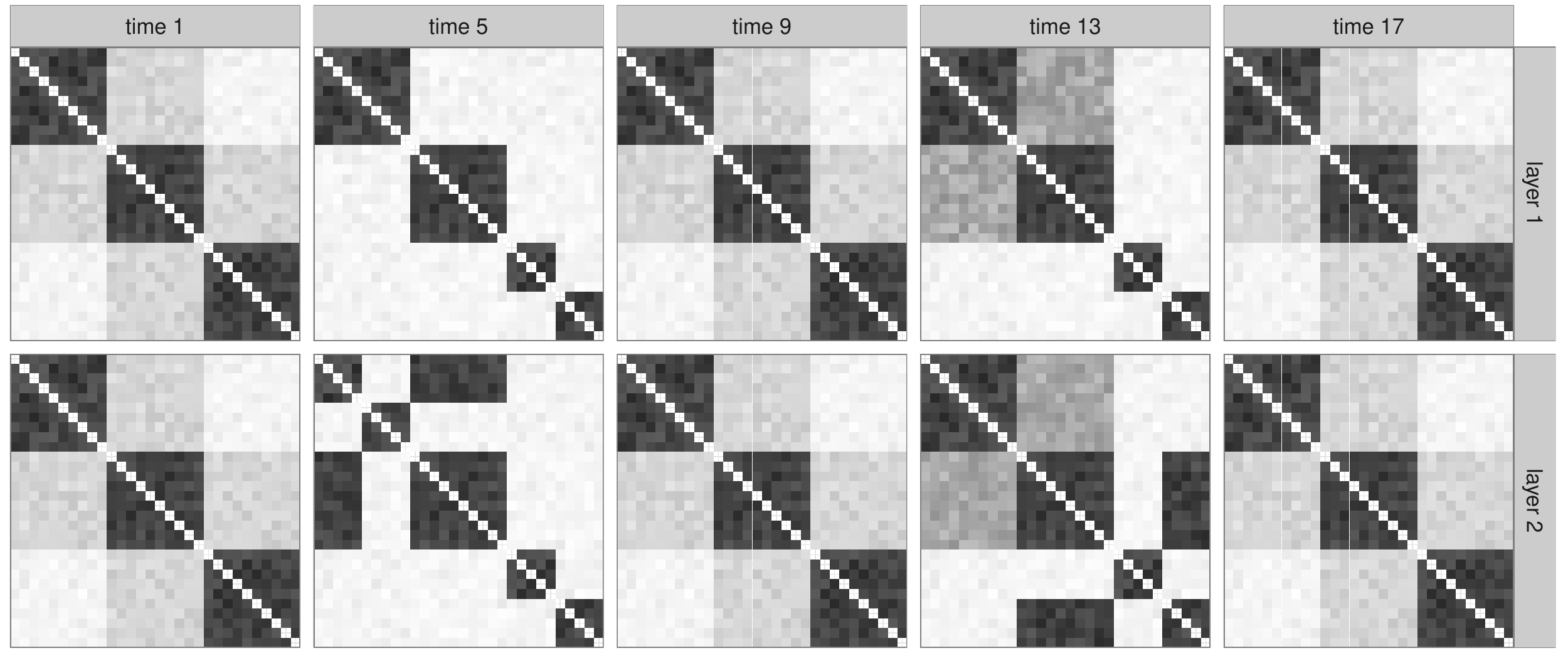}
\caption{True representative edge probability matrices $\pi^{0(k)}(t_i)$. The color in each cell goes from white to black as the corresponding edge probability goes from $0$ to $1$.}
\label{f2}
\end{figure}

We consider $V=30$ actors whose connections are monitored for $K=2$ days on the same time grid $t_i \in \mathbb{T}=\{1, \ldots, 17\}$. In simulating the dynamic multilayer networks, we sample the edges from conditionally independent Bernoulli variables given their corresponding edge probabilities as in equation \eqref{eq1}, with these probabilities defined to mimic possible scenarios in our application. To accomplish this, we consider, for each day, five representative edge probability matrices $\pi^{0(k)}(t_i)$ at times $t_{i}\in \{1,5,9,13,17\}$, displayed in Figure \ref{f2}, and define the matrices at the remaining times as a convex combination of these representative matrices. Focusing, for example, on the times $t_{14}$, $t_{15}$ and $t_{16}$ between $t_{13}$ and $t_{17}$, we let 
\begin{eqnarray*}
\pi^{0(k)}(t_{14})&=&0.75\pi^{0(k)}(t_{13}) + 0.25\pi^{0(k)}(t_{17}),\\
\pi^{0(k)}(t_{15})&=&0.50\pi^{0(k)}(t_{13}) + 0.50\pi^{0(k)}(t_{17}),\\
\pi^{0(k)}(t_{16})&=&0.25\pi^{0(k)}(t_{13}) + 0.75\pi^{0(k)}(t_{17}).
\end{eqnarray*}
The same construction holds for all the other time windows, except for times $t_4$ and $t_6$, where we induce a more rapid variation by letting  $\pi^{0(k)}(t_{4})=\pi^{0(k)}(t_{6})=\pi^{0(k)}(t_{5})$, instead of $\pi^{0(k)}(t_{4})=0.25\pi^{0(k)}(t_{1}) + 0.75\pi^{0(k)}(t_{5})$, and $\pi^{0(k)}(t_{6})=0.75\pi^{0(k)}(t_{5}) + 0.25\pi^{0(k)}(t_{9})$.  We incorporate this setting to study the effects of the Gaussian process' constant smoothness assumption in scenarios having time-varying smoothness. In order to empirically evaluate Proposition \ref{prop1}, the edge probability matrices in  Figure \ref{f2} are not generated from  equation \eqref{eq2}, but are instead constructed to characterize possible face-to-face contact scenarios.  

Focusing on the first day, which corresponds to layer $k=1$, the edge probability matrices at times $t_1$, $t_9$ and $t_{17}$ characterize contacts during breakfast, lunch and dinner, respectively.  These favor block-structures due to homophily by age and gender. In fact, it is reasonable to expect that young individuals, adult women and adult men, corresponding to actors in the first, second, and third block,  may have a higher chance of contact with individuals in the same socio-demographic group. We also allow adult women to have a moderate chance of contact with young individuals and adult men during breakfast, lunch, and dinner times.

The edge probability matrix at time $t_5$ characterizes block-structures due environmental restrictions during the morning, with  young individuals attending school, adult women sharing the house and two sub-groups of adult men working in two different places. Moreover, the edge probability matrix at time $t_{13}$ is similar to the one at time $t_{5}$ with exception of an increased chance of contact between  young individuals and adult women who are expected to share the same environment in the afternoon after school. 

The dynamic contact networks in the different days share similar patterns, as shown in Figure \ref{f1}. We maintain this property by considering the same representative  edge probability matrices in the second day, which corresponds to layer  $k=2$, with exception of $t_5$ and $t_{13}$. In the morning of the second day, we assume five young individuals contract a disease and remain at home. This increases their chance of contact  with the adult women taking care of them, and reduces the probability of a contact with the other young individuals. The edge probability matrix in the afternoon of the second day characterizes a similar scenario, but focuses on five adult men. Including different edge probability matrices at times $t_5$ and $t_{13}$, allows us to assess performance in learning layer-to-layer differences in contact patterns.

In order to assess predictive performance, we perform posterior analysis under our model and the competing methods, holding out from the observed data the networks from time $t_{13}$ to $t_{17}$ in the second day. This choice provides a scenario of interest in our application. For example a subset of actors may contract a disease at time $t_{12}$ in the second day, motivating the design of prevention policies relying on the forecasted contact patterns at future times.

\begin{table}[t]
\begin{center}
\def~{\hphantom{0}}
{%
\begin{tabular}{llllllllll}
 & \multicolumn{3}{l}{Squared Posterior Bias} &    \multicolumn{3}{l}{\ \ Posterior Variance \ \ \ \ } & \multicolumn{3}{l}{Posterior Concentration}\\ 
\hline 
 & & $n \downarrow$ & $V \downarrow$ &&  $n \downarrow$ &  $V \downarrow$  & &$n \downarrow$ & $V \downarrow$ \\
\hline
Joint & $ {\bf 0{.}009}$ & $0{.}014$ & $0{.}012$ & ${\bf 0{.}011}$  & $0{.}018$ & $0{.}016$  & ${\bf 0{.}020}$ & $0{.}032$ & $0{.}028$\\
Collapsed & $ {\bf 0{.}016}$ & $0{.}019$ & $0{.}019$ & ${\bf 0{.}007}$ & $0{.}011$ & $0{.}010$ & ${\bf 0{.}023}$  & $0{.}030$ & $0{.}029$\\
Separate & $ {\bf 0{.}016}$ & $0{.}025$ & $0{.}019$ & ${\bf 0{.}018}$ & $0{.}027$ & $0{.}023$ & ${\bf 0{.}034}$ & $0{.}052$ & $0{.}042$\\
 \hline
\end{tabular}}
\caption{For our methodology and the two  competitors, measures of posterior concentration around the true dynamic multilayer edge probability process, at varying $n$ and $V$. Bold numbers are the measures of concentration associated with our initial simulation scenario.}
\label{table1}
\end{center}
\end{table}

\subsection{Results in the Simulation}
\label{s52}
We perform posterior computation under our model with $\kappa_{\mu}=\kappa_{\bar{x}}=\kappa_{{x}}=0.05$ to favor smooth trajectories a priori and $a_1=2$, $a_2=2.5$ to facilitate  adaptation of the latent spaces dimensions. Although our model can be easily modified to learn these hyperparameters from the data as in \cite{mur_2010} and \cite{bhatt_2011}, respectively, borrowing of information within the dynamic multilayer networks has the effect of reducing sensitivity to the choice of the hyperparameters. In fact we found results robust to moderate changes in these quantities, and therefore prefer to elicit them to improve convergence and mixing. We consider $5000$ Gibbs iterations with $R=H=5$ and set a burn-in of $1000$. Traceplots for the time-varying edge probabilities in the two layers show no evidence against convergence, and mixing is good in our experience with most of the effective sample sizes for the quantities of interest being around 1500 out of 4000 samples. Posterior computation for our competitors is performed with the same smoothing parameters and $a_1=2$, $a_2=2.5$, using   $R+H=10$ dimensional latent spaces, to improve comparison with our results.

\subsubsection{Performance in Learning the True Edge Probability Process}
Flexible modeling of the dynamic multilayer edge probability process is  of paramount importance for accurate learning and prediction. Table \ref{table1} summarizes the concentration of the posterior distributions for the dynamic multilayer edge probabilities using the posterior mean of the squared difference between $\pi^{(k)}_{[vu]}(t_i) \mid Y_{\mbox{\footnotesize{obs}}}$ and $\pi^{0(k)}_{[vu]}(t_i)$. To provide empirical insights on posterior consistency we study this property also at varying $n$ and $V$. In the first case we focus on the dynamic multilayer network studied at odd times $t_i \in \{1,3,5, \ldots, 17\}$, whereas in the second case we consider the reduced networks where four actors within each block are not analyzed. To highlight the contribution of the squared posterior bias and the posterior variance, we factorize the concentration measure as
 $$\mbox{E}([\{\pi^{(k)}_{[vu]}(t_i) \mid Y_{\mbox{\footnotesize{obs}}}\}-\pi^{0(k)}_{[vu]}(t_i) ]^2)=[\mbox{E}\{\pi^{(k)}_{[vu]}(t_i)  \mid Y_{\mbox{\footnotesize{obs}}}\}-\pi^{0(k)}_{[vu]}(t_i) ]^2+\mbox{var}\{\pi^{(k)}_{[vu]}(t_i)  \mid Y_{\mbox{\footnotesize{obs}}}\}.$$ 
To assess the overall performance, these measures are averaged across the edge probabilities characterizing the dynamic multilayer network process.

According to Table \ref{table1}, borrowing of information across layers has the effect of reducing both bias and posterior variance compared to the separate analyses, improving accuracy and efficiency in inference and prediction. Enhancing flexibility via layer-specific latent coordinates additionally facilitates  bias reduction compared to collapsed analyses, but provides posterior distributions with slightly higher variance. This result is not surprising provided that the collapsed analyses assume the edge probability process to be common across layers, reducing the number of parameters. Moreover, although low variance is a desired feature when bias is small, this property may lead to  poor inference and prediction when the posterior distribution is centered far from the truth. In our simulation  --- characterized by  low separation between the true edge probability processes in the two layers  --- the bias is maintained at low levels also for the collapsed analyses, but this bias is expected to  increase when the layers have more evident differences.  

The above results are maintained at varying $n$ and $V$, while showing improved concentration for higher $n$ or $V$. Although it would be interesting to prove posterior consistency,  state-of-the-art literature provide results for  the single binary outcome case \citep[e.g][]{ghosal_2006}, and it is not clear how to adapt this theory to our much complex network--valued data. However, results in Table \ref{table1} provide positive empirical support for this property. 

\begin{figure}[t]
\centering
\includegraphics[width=14.5cm]{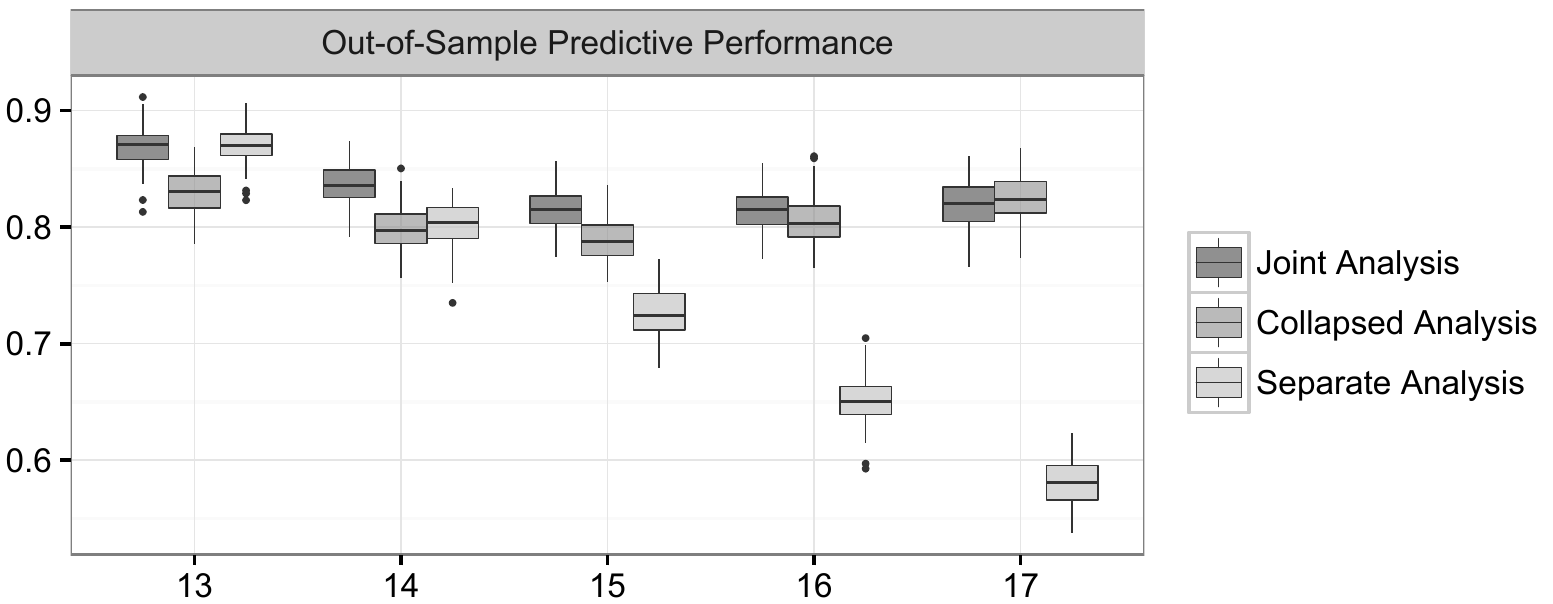}
\caption{For times from $t_{13}$ to $t_{17}$ --- held out in the second day from posterior computation --- box plots of the areas under the ROC curves (AUC) when predicting the unobserved edges with our method and the two competitors, under $100$ replicates of our simulated data. }
\label{f3}
\end{figure}

\subsubsection{Out-of-sample Predictive Performance}
As discussed in Sections \ref{s:2} and \ref{s:3}, bias and variance reduction are both fundamental tasks. However, developing statistical models characterized by lower bias is important to improve inference and prediction. In fact, our method has an improved ability to predict unobserved edges compared to the competing models, as shown in Figure \ref{f3}.\footnote{In assessing predictive performance in Figure \ref{f3}, we generate for each time from $t_{13}$ to $t_{17}$ --- held out in the second day  from posterior computation --- 100 networks from \eqref{eq1} with edge probabilities $\pi^{0(2)}_{[vu]}(t_i)$, and compute for each network the area under the ROC curve when predicting its edges  with the expectation of the posterior predictive distribution associated with each method --- according to Section \ref{sub:3}.}  In predicting held out edges we leverage the procedures in Section \ref{sub:3}. Since equation \eqref{eq1} is also valid for our competitors, results in Section \ref{sub:3} can be used also for the collapsed and separate analyses after replacing  $\mbox{E}\{\pi^{(k)}_{[vu]}(t_i) \mid Y_{\mbox{\footnotesize{obs}}}\}$ with $\mbox{E}\{\pi_{[vu]}(t_i) \mid Y_{\mbox{\footnotesize{obs}}}\}$ and $\mbox{E}\{\pi^{(k)}_{[vu]}(t_i) \mid Y^{(k)}_{\mbox{\footnotesize{obs}}}\}$, respectively.

\begin{figure}[t]
\centering
\includegraphics[width=13.7cm]{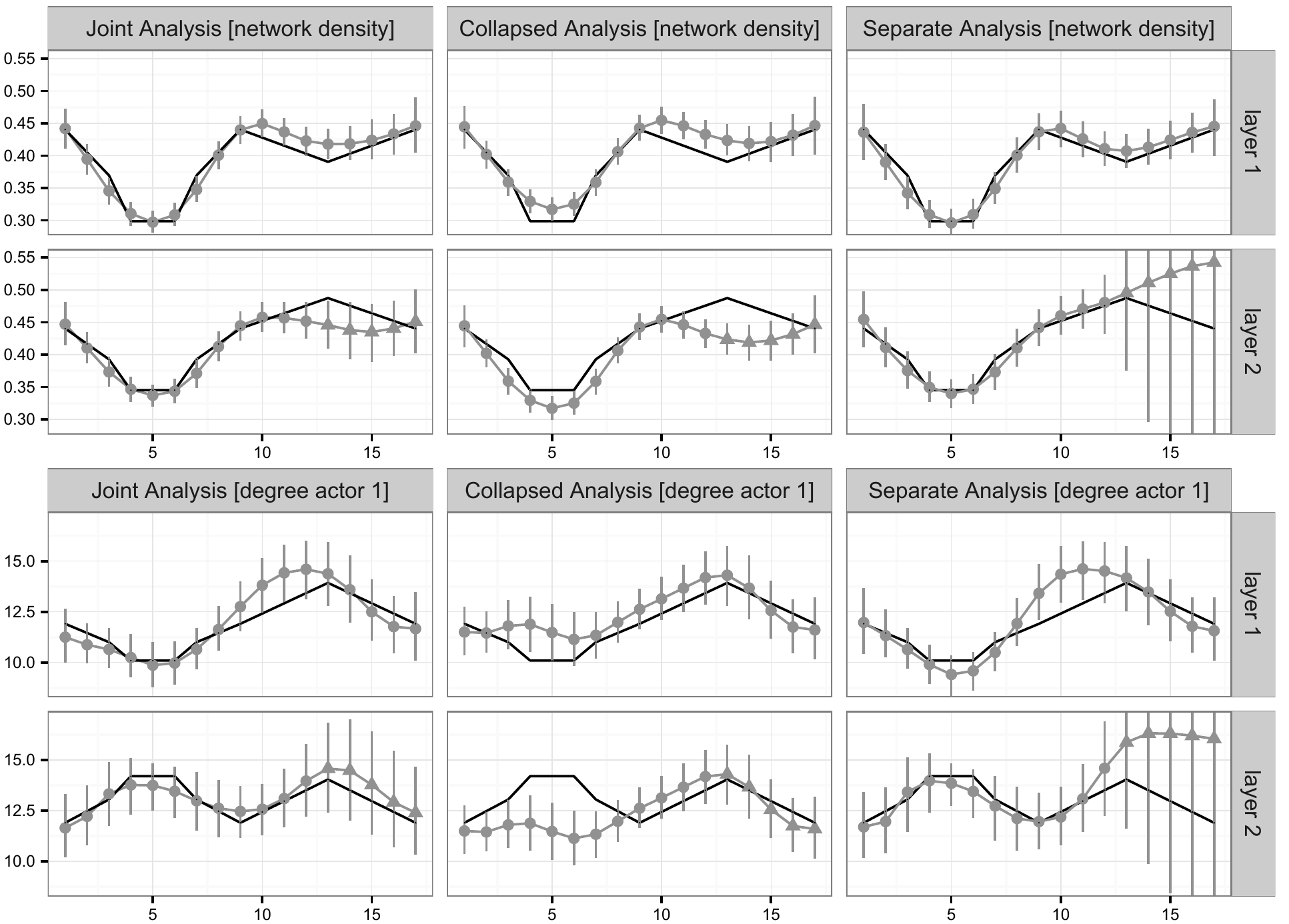}
\caption{For our method and the two competitors, posterior mean (grey lines) and point-wise $0.95$ credibility intervals (grey segments) for the dynamic expected network density (upper panels) and the dynamic expected degree of the first actor (lower panels) in the two layers. Black lines represent the true trajectories and triangles denote hold-out networks.}
\label{f4}
\end{figure}

According to Figure \ref{f3} the separate analysis obtains good predictive performance for the times subsequent to the last observed network, but this accuracy decays with increasing time lag. As the underlying edge probability process is smooth, it is reasonable to expect good predictive performance at time $t_{13}$ exploiting information at time $t_{12}$, whereas the reduced accuracy for later times is likely due to inefficient exploitation of the information from the first layer. As expected, the collapsed analysis provides accurate predictions when there is a high degree of similarity between the edge probabilities in the two layers, such as at time $t_{17}$, but the accuracy decreases in correspondence of more evident differences. Differently from the above procedures, our model incorporates dependence between layers without assuming a single edge probability process common to all of them, providing predictions which are less sensitive to the underlying data generating mechanism. In fact, as shown in Figure \ref{f3}, we obtain an  improved prediction of unobserved edges compared to the competing methods.

\subsubsection{Performance in Learning Network Properties of Interest}
Accurate modeling of the dynamic multilayer edge probabilities has key benefits in improving inference on time-varying network structures in the different layers. Figure \ref{f4} summarizes the posterior distribution of the dynamic expected network density and the time-varying expected degree of a selected actor in the two layers.

As shown in  Figure \ref{f4}, our method out-performs the competing approaches also in providing inference for the selected network properties. The collapsed analysis forces the trajectories of these networks summary measures to be equal across layers, causing poor performance in capturing differences among layers in specific times. Modeling the dynamic edge probability processes underlying each layer separately improves flexibility, but the inefficient exploitation of the information shared between layers leads to poor performance in modeling network properties associated with unobserved edges. Our formulation in equations \eqref{eq1}--\eqref{eq2} incorporates dependence between layers and preserves flexibility via a set of layer-specific latent coordinates, which reduces in-sample and out-of-sample bias. Although our model has accurate performance, the constant smoothness assumption characterizing the Gaussian process priors leads to a slight over-smoothing around the more rapid variation in times $t_4$ and $t_6$. One possibility to avoid this behavior is to replace the Gaussian process priors with nested Gaussian processes  \citep{zhu_2013} having time-varying smoothness.

\section{Application to Dynamic Multilayer Face-to-face Contact Data}
\label{s:5}
We now turn to applying the methods proposed in Sections \ref{s:2} and \ref{s:3} to the face-to-face contact networks described in Section \ref{s:1dat}. Raw contact data are available for $75$ individuals belonging to $5$ different households  in rural Kenya. Each household comprises different families living in the same compound and reporting to one head. Face-to-face proximity data for the individuals in each household are collected hourly from $07{:}00$ to $20{:}00$ during three consecutive days, with these windows of three days differing for most of the households. As a result, contact data are unavailable for almost all the actors in different households making it impossible for any learning methodology to infer across-households connectivity structures. In order to avoid complications due to non-overlapping days, we focus on the face-to-face  contact networks for the individuals in the most populated household comprising $V=29$ actors. This choice is further supported by the fact that within households contacts play a fundamental role in the speed of contagion,  motivating primary interests in household-based analyses; see e.g. \cite{house_2009} and the references cited therein.

\subsection{Results in the Application}
The face-to-face proximity data can be structured as a dynamic multilayer network recording for each day $k \in \{1, 2, 3\}$ the time-varying face-to-face contacts among the $V=29$ actors on the same hourly basis $t_1, \ldots, t_{14}$, where $t_i \in \{1, \ldots, 14\}$ is simply a discrete time index denoting the current hour. We apply our statistical model and the competing methods to the aforementioned data, where we hold out the contact network at the last time in the third day $Y_{t_{14}}^{(3)}$ to assess out-of-sample predictive performance.\footnote{Posterior computation under the three models uses the same default hyperparameters as in the simulation study. Similarly to the simulation study, we consider $5000$ Gibbs iterations and obtain no evidence against convergence after a burn-in of $1000$. The effective sample sizes for the quantities of interest for inference are around $1200$ out of $4000$ samples. Although this provides a slightly reduced result compared to the simulation studies, the mixing is still good given the complexity of our dynamic multilayer network data.}

\begin{table}[t]
\begin{center}
{%
\begin{tabular}{llccccccccccc}
 & &\multicolumn{5}{c}{Area under the curve}  & &  \multicolumn{5}{c}{Accuracy with cutoff 0.5} \\ 
\toprule
 \multicolumn{2}{l}{Number of latent coordinates} &&8& 10&  12 & &&&8&10&12&\\ 
 \midrule
Joint &In--sample& &0.97&0.97&0.97&&&&95\%&95\%&94\%&\\
 &Out--of--sample&& 0.95&0.95&0.95&&&&94\%&94\%&95\%&\\
 \cmidrule{2-13}
Collapsed  &In--sample&&0.94 & 0.94&0.95&&&&92\%&92\%&92\%&\\
&Out--of--sample&& 0.95&0.95&0.95&&&&94\%&94\%&94\%&\\
 \cmidrule{2-13}
Separate  & In--sample&&0.97&0.97&0.97&&&&94\%&94\%&94\%&\\
&Out--of--sample& &0.82&0.83&0.82&&&&89\%&89\%&89\%&\\
 \hline
\end{tabular}}
\caption{For our method and the two competitors, in-sample and out-of-sample predictive performance measured via the AUC, and the percentage of correctly predicted edges --- using a cutoff probability of $0.5$ --- at varying latent space dimensions $R+H$, with $R=H$. }
\label{f5}
\end{center}
\end{table}

\subsubsection{In-sample and Out-of-sample Predictive Performance}
As one assessment of our proposed methodology, Table \ref{f5} compares our performance for in-sample and out-of-sample edge prediction with those provided by the competing procedures. Consistent with the strategy outlined in equation  \eqref{eq9}, the ROC curves for in-sample edge prediction are constructed using the observed data $Y_{\mbox{\footnotesize{obs}}}$ and the posterior mean of their corresponding edge probabilities, estimated under the three different methods. The ROC curves for assessing out-of-sample predictive performance are defined in a similar way, but focusing on the held-out network at time $t_{14}$ in day $3$. To evaluate sensitivity to the choice of the upper bounds for the latent space dimensions, we study predictive performance also under other settings of $R$ and $H$, including $R=H=4$ and $R=H=6$. 

Edge prediction is clearly more accurate in-sample than out-of-sample. This is because $Y_{\mbox{\footnotesize{obs}}}$ is part of the posterior computation, whereas the network at time $t_{14}$ in day $3$ is held out. According to Table \ref{f5}, the separate analysis achieves comparable results to our model for in-sample edge prediction in accommodating layer-specific patterns, but provides poor predictions of future edges due to inability to incorporate information from the previous days. The collapsed analysis allows instead  borrowing of information between the three days via shared coordinates, but does not incorporate differences in contact patters across these days. As a result, the collapsed analysis provides comparable performance to our model for out-of-sample edge prediction, but has a reduced in-sample predictive ability. Our dynamic multilayer representation efficiently incorporates both shared structures and layer-specific differences, facilitating good predictive performance both in-sample and out-of-sample. As discussed in Section \ref{s:3}, these results  are not substantially affected by moderate variations in $R$ and $H$, and therefore we perform posterior inference for $R=H=5$, as in the simulation. 

\begin{figure}[t]
\centering
\includegraphics[width=13.8cm]{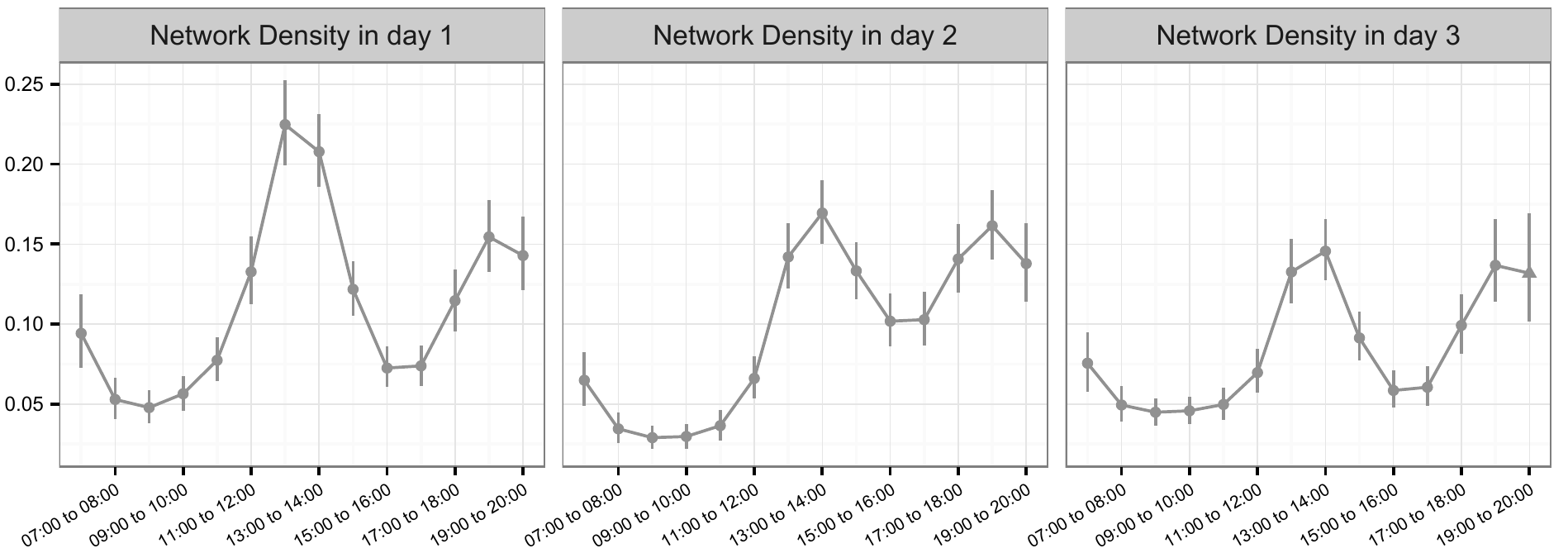}
\caption{Posterior mean (grey lines) and point-wise $0.95$ credibility intervals (grey segments) for the time-varying expected network density in the three days. The triangle for the last time in the third day denotes the contact network held out in posterior computation.}
\label{f6}
\end{figure}

\begin{figure}[t!]
\centering
\includegraphics[width=14.8cm]{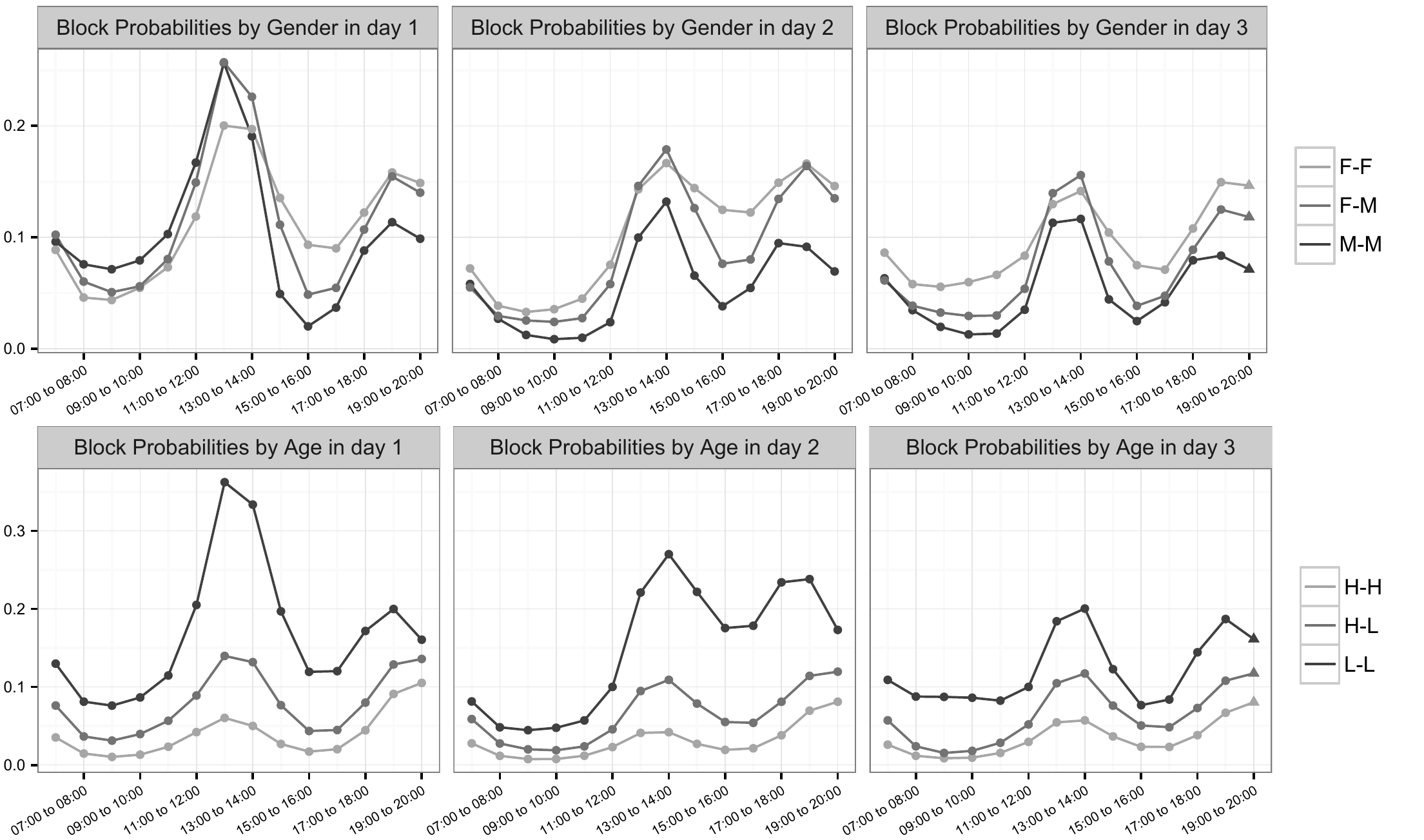}
\caption{Upper panels: posterior mean of the dynamic expected frequencies of contact for actors with the same gender --- both males or both females --- and different gender,  in each day. Lower panels: same quantities with respect to high ($\geq 15$) and low  ($<$15) age groups.}
\label{f7}
\end{figure}

\subsubsection{Learning Network Properties of Interest}
The above results motivate additional posterior analyses of the dynamic multilayer stochastic process estimated under our model. Figure \ref{f6} summarizes the posterior distribution for the trajectory of the expected network density in the three days. The expected frequency of proximity contacts evolves on similar patterns during the three days, while showing some day-specific variations at certain hours. This result further confirms the importance of analyzing such data with statistical models that are able to borrow information across days, while maintaining flexibility in capturing day-specific patterns. Consistent with  \cite{kiti_2016}, the expected network density remains in general on low levels during the morning and afternoon when the actors are in different environments such as school and workplace, and peaks  in correspondence of lunch and dinner times when the households members congregate. We learn a similar pattern when considering the dynamic expected frequencies of contact within and between groups of actors having similar traits, such as gender and age.\footnote{We can easily derive the posterior mean of these trajectories by averaging, for each time $t_{i}$ and day $k$, all the estimated edge probabilities corresponding to pairs of actors having the same combination of traits, e.g. both males, both females, one male and one female.} These trajectories provide key additional information compared to the expected network density, highlighting how infectious diseases can spread within and between groups of actors. The posterior mean of these trajectories is visualized in Figure \ref{f7}. 

\begin{figure}[t]
\centering
\includegraphics[width=15.2cm]{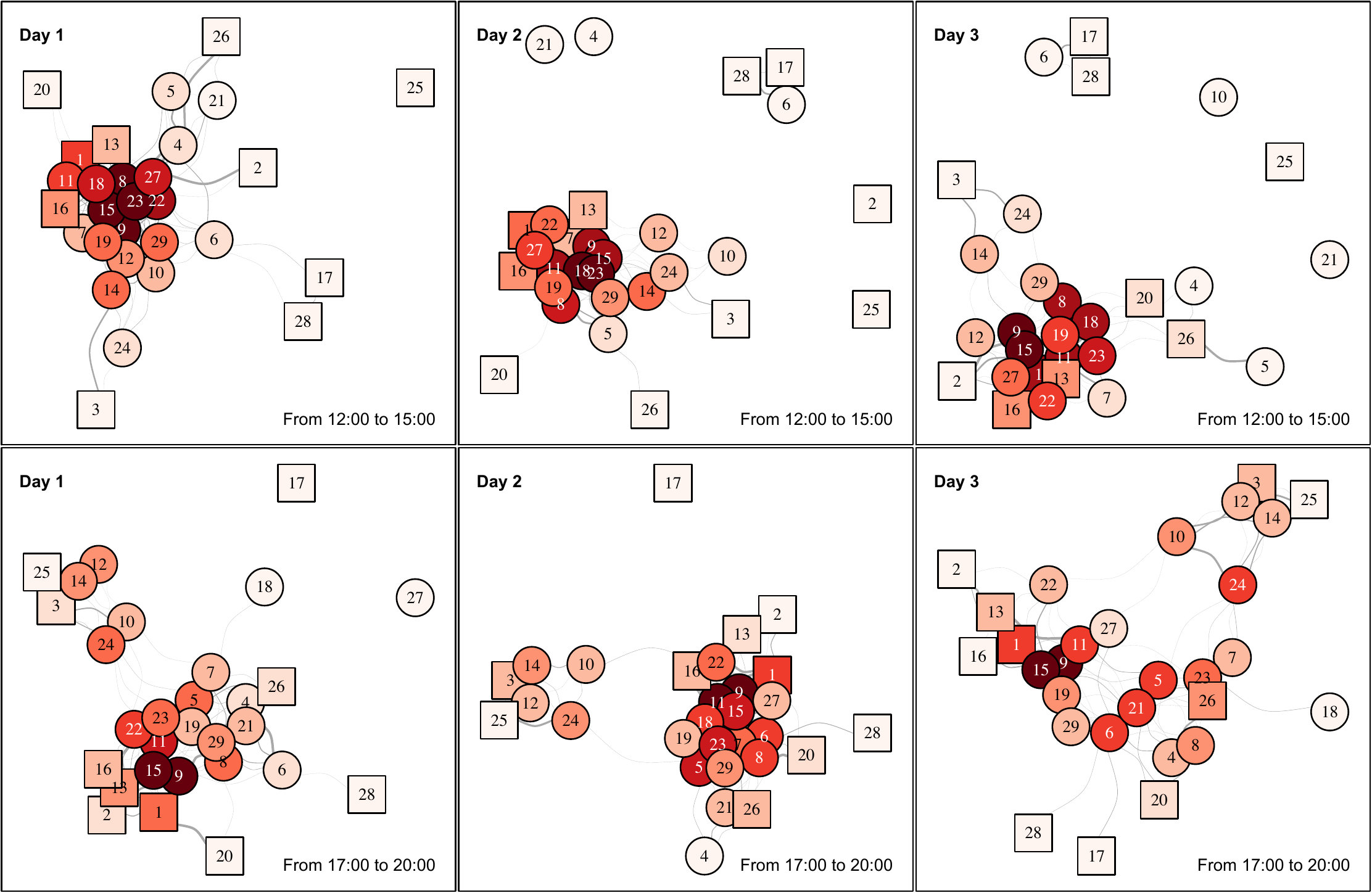}
\vspace{-10pt}
\caption{Weighted network visualization with weights obtained by averaging the posterior mean of the edge probabilities for the two time windows having the more dense networks, within each of the three days. Squares and circles represent the actors having high ($\geq 15$) and low ($<15$) age, respectively. The color of each node goes from white to dark red as the degree of the corresponding actor increases, relative to the others. Node positions are obtained by applying the \citet{fru_1991} force--directed placement algorithm, whereas the width of the each edge is proportional to its averaged edge probability.}
\label{f8}
\end{figure}

According to Figure \ref{f7}, all the trajectories inherit the pattern discussed for the expected network density, with clear peaks during lunch and dinner times. Although there are not evident differences in the dynamic expected frequencies of face-to-face contact within and between gender groups, we  notice how contacts among males are typically less likely than contacts between females, whereas face-to-face interactions among actors of different genders become notable during lunch. Therefore, these hours might be at risk of spreading diseases from a gender group to the other. Consistent with the static analyses considered in \cite{kiti_2016}, we observe more differences in the trajectories by age groups, with a substantially high chance of contact between young actors, whereas the dynamic expected frequency of interaction among adults remains on low levels. In fact, young individuals potentially have less restrictions from environment or work schedule than adults and therefore have more chances of interaction, especially with actors of their same age. Hence, diseases may spread more easily among young individuals and reach adults during lunch times when the chance of contact between the two age groups is higher. 

Figures \ref{f6} and \ref{f7} provide global measures for the dynamic risk of contagion in the network and for groups of actors. In order to develop more refined prevention policies it is of key interest to monitor the  infectivity of each actor and how this measure evolves with time and days. Consistent with this goal, Figure \ref{f8} provides a graphical analysis of selected contact networks, with the actors positions depending on the estimated edge probabilities averaged, for each day, over the two time windows having the more dense networks which may lead to a faster disease spread. As shown in Figure \ref{f8},  the contact patterns in the same time window share a similar configuration across the three days, while displaying some day-to-day difference in the connectivity behavior of  subsets of actors. For example, the community structure formed by actors $6$, $17$ and $28$ during lunch times is less separated from the rest of the network in the first day, compared to the second and third day. This result provides additional support for our methodology which can simultaneously incorporate shared patterns and possible day-specific deviations from these common structures. Consistent with Figures \ref{f6} and \ref{f7}, young individuals --- represented by circles --- are typically characterized by more frequent contacts, whereas the adult actors --- corresponding to squares --- have a more peripheral position in the network. As a result the degree of the young individuals is typically higher, making these actors  particularly relevant in the economy of disease contagion and transmission. In this respect, actors 9 and 15 may represent dangerous hubs. 

We can additionally notice some community structure within the network, particularly during dinner times. For example, actors 3, 10, 12, 14, 24 and 25 are tightly connected, but separated from the others, with exception of the young subjects 10 and 24 which may play a key role in the transmission of a disease from their community to the rest of the network. Other two communities can be more clearly observed during dinner times in day three. It is interesting to notice how these groups of actors comprise both young and adult individuals, and therefore may represent different families within the household under analysis.

\section{Discussion}
\label{s:6}
The increasing availability of multidimensional, complex, and dynamic information on  social interaction processes, motivates a growing demand for novel statistical models. In order to successfully enhance quality in inference and prediction, these models need to efficiently incorporate the complex set of dependencies in the observed data, without affecting flexibility.  Motivated by this consideration and by epidemiological studies monitoring disease transmission via dynamic face-to-face interactions, we have developed a Bayesian nonparametric model for dynamic multilayer networks leveraging latent space representations. In order to preserve flexibility and borrow information across layers and time, we modeled the edge probabilities as a function of shared and layer-specific latent coordinates which evolve in time via Gaussian process priors. We provided theoretical support for our model and developed simple procedures for posterior computation and formal prediction. Finally, we illustrated on both simulated data and on infection studies monitoring face-to-face contacts that our methods perform better than competitors in terms of inference and prediction.

Although we focus on face-to-face interaction networks collected at multiple times and days, our methods have a broad range of applications. Notable examples include dynamic cooperations among countries with respect to different types of international relations, time-varying interactions between researchers according to multiple forms of academic collaborations and dynamic contacts between terrorists in relation to different types of dark interactions. In all these relevant applications, our flexible methodology can provide an appealing direction in accurately learning and predicting hidden wiring mechanisms and their implication in several environments and phenomena.

In addition, our methods motivate further directions of research. An important one is
to facilitate scaling to larger dynamic multilayer network data. Currently, the computational complexity of our Gibbs sampler --- as a function of the dimensions of the input data --- is of order  $O(VKn^3)$, and corresponds to the most intensive step updating the layer-specific latent coordinates trajectories for the $V$ actors. Although the latent space representation reduces computational complexity from quadratic in the number of actors $V$ to linear, the cubic complexity in the number of time points $n$ associated with the Gaussian process prior, may still represent a computational barrier when data are monitored for wide time windows.  A possible strategy to successfully address this issue is to consider more scalable processes such as the low-rank approximations to the Gaussian process \citep{baner_2012} or state-space models. Another important generalization is accommodating the more informative contact counts instead of just a binary variable indicating presence or absence of face-to-face interactions. In accomplishing this goal one possibility is to adapt the methodology proposed in  \cite{can_2011} to our framework and assume the weighted edges are realizations from a rounded Gaussian whose mean is factorized as in equation \eqref{eq2}.

\section*{Appendix A: Proofs of the Propositions \ref{prop1} and \ref{prop2}}
{\bf Proof of Proposition \ref{prop1}}. As the logistic mapping is one-to-one continuous, Proposition \ref{prop1} is valid if and only if --- for each time $t \in \mathbb{T}$ --- every possible collection of log-odds $z(t)=\{ z^{(k)}_{[vu]}(t) \in \Re: k=1, \ldots, K, \ v=2, \ldots, V,\ u=1, \ldots, v-1\}$ can be factorized as $$z^{(k)}_{[vu]}(t)=\mu(t)+\bar{s}_{[vu]}(t)+{s}^{(k)}_{[vu]}(t)=\mu(t)+\bar{x}_{v}(t)^{\T}\bar{x}_{u}(t)+{x}^{(k)}_{v}(t)^{\T}{x}^{(k)}_{u}(t),$$
for all $k=1, \ldots, K, \ v=2, \ldots, V,\ u=1, \ldots, v-1$.

Assume without loss of generality $\mu(t)=0$ for every $t \in \mathbb{T}$ and let $\bar{X}(t)$ and ${X}^{(k)}(t)$ for $k=1, \ldots, K$, denote the $V \times R$ and $V \times H$ matrices containing the shared and layer-specific latent coordinates, respectively, at time $t \in \mathbb{T}$. Since we are not interested in modeling the diagonal elements of the $V \times V$ edge probabilities and log-odds matrices, it is always possible to write
$$Z^{(k)}(t)=\bar{S}(t)+S^{(k)}(t), \quad k=1, \ldots, K,$$
where $Z^{(k)}(t)$ is the $V \times V$ symmetric matrix having the log-odds of the edges at time $t$ in layer $k$ as off-diagonal elements, whereas $\bar{S}(t)$ and $S^{(k)}(t)$ are $V \times V$ positive semidefinite symmetric matrices having quantities $\bar{s}_{[vu]}(t)$, $v=2, \ldots, V,\ u=1, \ldots, v-1$ and  $s^{(k)}_{[vu]}(t)$, $v=2, \ldots, V,\ u=1, \ldots, v-1$ as off-diagonal elements. Since we are not interested in self-relations, there is no loss of generality in assuming $\bar{S}(t)$ and $S^{(k)}(t)$ positive semidefinite, since for any configuration of shared and layer-specific similarities there exist infinitely many positive semidefinite matrices having these quantities as off-diagonal elements. 

Since $\bar{S}(t)$ and $S^{(k)}(t)$ are positive semidefinite, they admit the eigen-decompositions $\bar{S}(t)=\bar{U}(t)\bar{\Lambda}(t) \bar{U}(t)^{\T}$ and ${S}^{(k)}(t)={U}^{(k)}(t)\Lambda^{(k)}(t) {U}^{(k)}(t)^{\T}$, where $\bar{U}(t)$ and ${U}^{(k)}(t)$ denote the $V \times {R}_t$ and $V \times {H}_t^{(k)}$ matrices of eigenvectors, whereas $\bar{\Lambda}(t)$ and $\Lambda^{(k)}(t)$ are the corresponding diagonal matrices with the positive eigenvalues. Therefore, letting $R \geq {R}_t$ for every $t \in \mathbb{T}$ and $H \geq {H}_t^{(k)}$, for every $k=1, \ldots, K$ and $t \in \mathbb{T}$, Proposition \ref{prop1} follows after defining $\bar{X}(t)$ and ${X}^{(k)}(t)$ as the block matrices $\bar{X}(t)=\{\bar{U}(t)\bar{\Lambda}(t)^{1/2}, 0_{V \times (R-{R}_t)} \}$ and ${X}^{(k)}(t)=\{{U}^{(k)}(t){\Lambda}^{(k)}(t)^{1/2}, 0_{V \times (H-{H}^{(k)}_t)} \}$ for every $k=1, \ldots, K$.  $\Box$

\vspace{20pt}

\noindent {\bf Proof of Proposition \ref{prop2}}. Leveraging proof of Corollary 2 in \cite{dur_2014}, to prove Proposition \ref{prop2} it suffices to show that 
\begin{eqnarray}
\mathrm{pr}\left(\mathrm{sup}_{t\in \mathbb{T}} \left[\sum_{k=1}^{K}\sqrt{ \sum_{v=2}^V \sum_{u=1}^{v-1} \{ z^{(k)}_{[vu]}(t) -z^{0(k)}_{[vu]}(t) \}^2}\right]<\epsilon \right)>0,
\label{eq10}
\end{eqnarray}
where $z^{(k)}_{[vu]}(t)$ and $z^{0(k)}_{[vu]}(t)$ are the log-odds of $\pi^{(k)}_{[vu]}(t)$ and $\pi^{0(k)}_{[vu]}(t)$, respectively. Recalling the proof of Proposition \ref{prop1}, the above probability can be factorized as
$$\mathrm{pr}\left(\mathrm{sup}_{t\in \mathbb{T}} \left[\sum_{k=1}^{K}\sqrt{ \sum_{v=2}^V \sum_{u=1}^{v-1} \{ \mu(t)+\bar{s}_{[vu]}(t)+{s}^{(k)}_{[vu]}(t) -\mu^0(t)-\bar{s}^0_{[vu]}(t)-{s}^{0(k)}_{[vu]}(t) \}^2}\right]<\epsilon \right),$$
with $\bar{s}_{[vu]}(t)=\bar{x}_{v}(t)^{\T}\bar{x}_{u}(t)$, ${s}^{(k)}_{[vu]}(t)={x}^{(k)}_{v}(t)^{\T}{x}^{(k)}_{u}(t)$, $\bar{s}^0_{[vu]}(t)=\bar{x}^{0}_{v}(t)^{\T}\bar{x}^{0}_{u}(t)$ and ${s}^{0(k)}_{[vu]}(t)={x}^{0(k)}_{v}(t)^{\T}{x}^{0(k)}_{u}(t)$. Exploiting the triangle inequality and the independence of the Gaussian process priors for the different trajectories, a lower bound for the above probability is
\begin{eqnarray*}
\mathrm{pr}\left(\mathrm{sup}_{t\in \mathbb{T}} \left[\sqrt{ \sum_{v=2}^V \sum_{u=1}^{v-1} \{ \mu(t) -\mu^0(t)+\bar{s}_{[vu]}(t)-\bar{s}^0_{[vu]}(t)\}^2}\right]<\frac{\epsilon}{2K} \right) \times \\ \times \prod_{k=1}^K\mathrm{pr}\left(\mathrm{sup}_{t\in \mathbb{T}} \left[\sqrt{ \sum_{v=2}^V \sum_{u=1}^{v-1} \{ {s}^{(k)}_{[vu]}(t) -{s}^{0(k)}_{[vu]}(t)\}^2}\right]<\frac{\epsilon}{2K} \right). \ \ \ \ \ \ \  \ \ \ \ \ \ \ \ \ \
\end{eqnarray*}
Applying Theorem 2 in \cite{dur_2014} to each term in this factorization it is easy to show that all the above probabilities are strictly positive, proving Proposition \ref{prop2}. $\Box$

\section*{Appendix B: Pseudo-Code for Posterior Computation}
Algorithm \ref{Algorithm_1} provides guidelines for step-by-step implementation of our Gibbs sampler.
\begin{breakablealgorithm}
\caption{Gibbs sampler for the dynamic multilayer latent space model}
\begin{algorithmic}
\STATE  {{\bf [1] Generate the P\'olya-gamma augmented data}}
\FOR{each $t_i=t_1, \ldots, t_n$, $k=1, \ldots K$, $v=2, \ldots, V$ and $u=1, \ldots, v-1$}
\STATE Sample the augmented data $\omega^{(k)}_{[vu]}(t_i)$ from the full conditional P\'olya-gamma 
\begin{eqnarray*}
\ \ \omega^{(k)}_{[vu]}(t_i) \mid - \sim \mbox{\small{PG}}\left\{1,\mu(t_i)+\bar{x}_{v}(t_i)^{\T}\bar{x}_{u}(t_i)+{x}^{(k)}_{v}(t_i)^{\T}{x}^{(k)}_{u}(t_i)\right\},
\end{eqnarray*}
where $\mbox{\small{PG}}(a,b)$ is the P\'olya-gamma random variable with parameters $a>0$ and $b \in \Re$.
\ENDFOR
\STATE ---------------------------------------------------------------------------------------------------------------------
\STATE {{\bf  [2] Update the baseline trajectory $\mu=\{\mu(t_1),\ldots,\mu(t_n)\}^{\T}$}} from $\mu \mid -   \sim \mbox{N}_{n}(\mu_\mu, \Psi_{\mu})$
where $\Psi_{\mu}=\left[\mbox{diag}\left\{\sum_{k=1}^{K}\sum_{v=2}^{V}\sum_{u=1}^{v-1} \omega^{(k)}_{[vu]}(t_1),\dots,\sum_{k=1}^{K}\sum_{v=2}^{V}\sum_{u=1}^{v-1} \omega^{(k)}_{[vu]}(t_n) \right\}+ \Sigma_{\mu}^{-1}\right]^{-1}$ and $\mu_\mu=\Psi_{\mu} \eta_{\mu}$ with 
\begin{eqnarray*}
 \eta_{\mu}=\left[ \begin{array}{c}
\sum_{k=1}^{K}\sum_{v=2}^{V}\sum_{u=1}^{v-1}\{Y^{(k)}_{t_1[vu]}-1/2- \omega^{(k)}_{[vu]}(t_1)[\bar{x}_{v}(t_1)^{\T}\bar{x}_{u}(t_1)+{x}^{(k)}_{v}(t_1)^{\T}{x}^{(k)}_{u}(t_1)]\}\\
\vdots\\
\sum_{k=1}^{K}\sum_{v=2}^{V}\sum_{u=1}^{v-1}\{Y^{(k)}_{t_n[vu]}-1/2- \omega^{(k)}_{[vu]}(t_n)[\bar{x}_{v}(t_n)^{\T}\bar{x}_{u}(t_n)+{x}^{(k)}_{v}(t_n)^{\T}{x}^{(k)}_{u}(t_n)]\}
 \end{array} \right].
\end{eqnarray*}
\vspace{-5pt}
\STATE ---------------------------------------------------------------------------------------------------------------------
\STATE {{\bf [3] Sample the vectors of shared coordinates $\bar{x}_{v}(t_1), \ldots, \bar{x}_{v}(t_n)$ for $v=1, \ldots, V$}}
\FOR{each actor $v=1, \ldots, V$}
\STATE Block-sample $\{\bar{x}_{v}(t_1), \ldots, \bar{x}_{v}(t_n)\}$ given the others $\{\bar{x}_{u}(t_i): u \neq v, t_i=t_1, \ldots t_n \}$. 
\begin{description}
\item{[a] Let $\bar{x}_{(v)}=\{\bar{x}_{v1}(t_1),\ldots,\bar{x}_{v1}(t_n), \dots,\bar{x}_{vR}(t_1),\ldots,\bar{x}_{vR}(t_n) \}^{\T}$}
\item{[b] Define a Bayesian logistic regression with $\bar{x}_{(v)}$ acting as coefficient vector and  having prior, according to equation \eqref{eq7}, $\bar{x}_{(v)}\sim \mbox{N}_{n\times R}\left\{ 0,\mbox{diag}(\tau^{-1}_1,\ldots,\tau^{-1}_R) \otimes \Sigma_{\bar{x}} \right\}$}
\item{[c] For every $k=1, \ldots, K$, the Bayesian logistic regression for the updating of $\bar{x}_{(v)}$ is 
\begin{eqnarray*}
{Y}^{(k)}_{(v)}  \sim \mbox{Bern}(\pi^{(k)}_{(v)}), \quad \mbox{logit}(\pi^{(k)}_{(v)})=1_{V-1}\otimes \mu +\bar{X}_{(-v)}\bar{x}_{(v)}+X^{(k)}_{(-v)}x^{(k)}_{(v)},
\end{eqnarray*}
with
\vspace{-7pt}
\begin{itemize}
\item{${Y}^{(k)}_{(v)} $ obtained by stacking vectors $\{{Y}^{(k)}_{t_1[vu]}, \ldots, {Y}^{(k)}_{t_n[vu]}\}^\T$ for all  pairs having $v$ as a one of the two nodes.}
\vspace{-7pt}
\item{$\pi^{(k)}_{(v)}$ the corresponding vector of  edge probabilities}
\vspace{-7pt}
\item{$x^{(k)}_{(v)}=\{{x}^{(k)}_{v1}(t_1),\ldots,{x}^{(k)}_{v1}(t_n), \dots, {x}^{(k)}_{vH}(t_1),\ldots, {x}^{(k)}_{vH}(t_n) \}^{\T}$}
\vspace{-7pt}
\item{$\bar{X}_{(-v)}$ and ${X}^{(k)}_{(-v)}$the matrices of regressors whose entries are suitably chosen from  $\{\bar{x}_{u}(t_i): u \neq v, t_i=t_1, \ldots t_n \}$ and $\{{x}^{(k)}_{u}(t_i): u \neq v, t_i=t_1, \ldots t_n \}$}
\end{itemize}
\vspace{-7.5pt}
According to the above specification and letting  $\Omega^{(k)}_{(v)}$ the diagonal matrix with the corresponding P\'olya-gamma augmented data, we obtain
\begin{eqnarray*}
\bar{x}_{(v)}\mid - \sim \mbox{N}_{n\times R}\left( \mu_{\bar{x}_{(v)}},\Psi_{\bar{x}_{(v)}} \right),
\end{eqnarray*}
with 
\begin{description}
\item{$\Psi_{\bar{x}_{(v)}}=\{ \bar{X}^{\T}_{(-v)}(\Omega^{(1)}_{(v)}+\ldots+\Omega^{(K)}_{(v)})\bar{X}_{(-v)} +\mbox{diag}(\tau_1,\ldots,\tau_R)\otimes \Sigma_{\bar{x}}^{-1}  \}^{-1}$}
\vspace{-7.5pt}
\item{$ \mu_{\bar{x}_{(v)}}=\Psi_{\bar{x}_{(v)}}( \bar{X}^{\T}_{(-v)} [\sum_{k=1}^K\{{Y}^{(k)}_{(v)}- \ 1_{V-1}\otimes1_{n}0.5  -\Omega^{(k)}_{(v)}(1_{V-1}\otimes  \mu+X^{(k)}_{(-v)}x^{(k)}_{(v)})\}])$}
\end{description}
}
\end{description}
\ENDFOR
\STATE ---------------------------------------------------------------------------------------------------------------------
\STATE {{\bf [4] Sample the layer-specific coordinates ${x}^{(k)}_{v}(t_1), \ldots, {x}^{(k)}_{v}(t_n)$ for $v=1, \ldots, V$ and $k=1, \ldots, K$}}
\FOR{each layer $k=1, \ldots, K$}
\FOR{each actor $v=1, \ldots, V$}
\STATE Block-sample $\{{x}^{(k)}_{v}(t_1), \ldots, {x}^{(k)}_{v}(t_n)\}$ given  $\{{x}^{(k)}_{u}(t_i): u \neq v, t_i=t_1, \ldots t_n \}$. In particular, letting ${x}^{(k)}_{(v)}=\{{x}^{(k)}_{v1}(t_1),\ldots,{x}^{(k)}_{v1}(t_n), \dots,{x}^{(k)}_{vH}(t_1),\ldots,{x}^{(k)}_{vH}(t_n) \}^{\T}$ and adapting derivations in step [3] to the sampling of  the layer-specific coordinates, it is straightforward to obtain \begin{eqnarray*}
{x}^{(k)}_{(v)}\mid - \sim \mbox{N}_{n\times H}\left( \mu_{{x}^{(k)}_{(v)}},\Psi_{{x}^{(k)}_{(v)}} \right),
\end{eqnarray*}
with 
\vspace{-7.5pt}
\begin{description}
\item{$\Psi_{{x}^{(k)}_{(v)}}=\{ {X}_{(-v)}^{(k)\T}\Omega^{(k)}_{(v)}{X}^{(k)}_{(-v)} +\mbox{diag}(\tau^{(k)}_1,\ldots,\tau^{(k)}_H)\otimes \Sigma_{{x}}^{-1}  \}^{-1}$}
\vspace{-7.5pt}
\item{$ \mu_{{x}^{(k)}_{(v)}}=\Psi_{{x}^{(k)}_{(v)}}[{X}_{(-v)}^{(k)\T} \{{Y}^{(k)}_{(v)}- \ 1_{V-1}\otimes1_{n}0.5  -\Omega^{(k)}_{(v)}(1_{V-1}\otimes  \mu+\bar{X}_{(-v)}\bar{x}_{(v)})\}]$}
\end{description}

\ENDFOR
\ENDFOR
\STATE ---------------------------------------------------------------------------------------------------------------------
\STATE {{\bf [5] Update the gamma quantities defining the shared shrinkage parameters $\tau^{-1}_1, \ldots, \tau^{-1}_R$}}
\begin{eqnarray*}
\delta_{1} \mid - &\sim& \mbox{Ga} \left\{a_{1}+\frac{V\times n \times R}{2},1+\frac{1}{2}\sum_{m=1}^{R}\theta^{(-1)}_m\sum_{v=1}^{V}\bar{x}_{vm}^{\T} \Sigma_{\bar{x}}^{-1}\bar{x}_{vm}\right\}, \quad \quad \quad \quad \nonumber \\
\delta_{r \geq 2} \mid - &\sim& \mbox{Ga}\left\{a_{2}+\frac{V\times n\times (R-r+1)}{2},1+\frac{1}{2}\sum_{m=r}^{R}\theta^{(-r)}_m\sum_{v=1}^{V}\bar{x}_{vm}^{\T} \Sigma_{\bar{x}}^{-1}\bar{x}_{vm}\right\},
\end{eqnarray*}
where $\theta_{m}^{(-r)}=\prod_{t=1,t\neq r}^{m} \delta_{t}$ for $r=1,\ldots,R$ and $\bar{x}_{vm}=\{\bar{x}_{vm}(t_1), \dots, \bar{x}_{vm}(t_n)\}^{\T}$.
\STATE ---------------------------------------------------------------------------------------------------------------------
\STATE {{\bf [6] Sample the variables defining the layer-specific parameters $\tau^{(k)-1}_1, \ldots, \tau^{(k)-1}_H$, $k=1, \ldots, K$}}
\FOR{each layer $k=1, \ldots, K$}
\STATE 
\vspace{-15pt}
\begin{eqnarray*}
\delta^{(k)}_{1} \mid - &\sim& \mbox{Ga} \left\{a_{1}+\frac{V\times n \times H}{2},1+\frac{1}{2}\sum_{l=1}^{H}\theta^{(-1)}_l\sum_{v=1}^{V}{x}_{vl}^{\T} \Sigma_{{x}}^{-1}{x}_{vl}\right\}, \quad \quad \quad \quad \nonumber \\
\delta^{(k)}_{h \geq 2} \mid - &\sim& \mbox{Ga}\left\{a_{2}+\frac{V\times n\times (H-h+1)}{2},1+\frac{1}{2}\sum_{l=h}^{H}\theta^{(-h)}_l\sum_{v=1}^{V}{x}_{vl}^{\T} \Sigma_{{x}}^{-1}{x}_{vl}\right\},
\end{eqnarray*}
where $\theta_{l}^{(-h)}=\prod_{t=1,t\neq h}^{l} \delta^{(k)}_{t}$ for $h=1,\ldots,H$ and ${x}_{vl}=\{{x}^{(k)}_{vl}(t_1), \dots, {x}^{(k)}_{vl}(t_n)\}^{\T}$.\ENDFOR
\STATE ---------------------------------------------------------------------------------------------------------------------
\STATE {{\bf [7] Obtain the posterior samples for the edge probabilities}} $\pi^{(k)}_{[vu]}(t_i)$ via $$\pi^{(k)}_{[vu]}(t_i)=[1+\exp\{-\mu(t_i)-\bar{x}_{v}(t_i)^{\T}\bar{x}_{u}(t_i)-{x}^{(k)}_{v}(t_i)^{\T}{x}^{(k)}_{u}(t_i)\}]^{-1}$$ for every $t_i=t_1, \ldots, t_n$, layer $k=1, \ldots, K$ and actors $v=2, \ldots, V$, $u=1, \ldots, v-1$. 
\end{algorithmic}
\label{Algorithm_1}
\end{breakablealgorithm}

\bibliography{mybibfile}

\end{document}